\documentclass{article}
\usepackage{arxiv}
\usepackage[utf8]{inputenc} 
\usepackage[T1]{fontenc}    
\usepackage[hyperfootnotes=false]{hyperref}       
\usepackage{url}            
\usepackage{booktabs}       
\usepackage{amsfonts}       
\usepackage{nicefrac}       
\usepackage{microtype}      
\usepackage{lipsum}		
\usepackage{graphicx}
\usepackage{grffile}
\usepackage[numbers]{natbib}
\usepackage{doi}
\usepackage{amsmath}
\usepackage{amsthm}
\usepackage{subcaption}
\usepackage{footnote}
\makesavenoteenv{table*}
\makesavenoteenv{tabular}
\usepackage{tablefootnote}
\usepackage{mwe}
\usepackage{float}
\usepackage{ragged2e}
\usepackage{multirow}
\usepackage{wrapfig}
\usepackage{xcolor}
\floatstyle{plaintop}
\restylefloat{table}

\newcommand*{\slashn}[1]{\texttt{\textbackslash#1}}

\usepackage{etoolbox}

\title{Kandinsky 3.0 Technical Report}

\author{{\hspace{1mm}Vladimir Arkhipkin$^1$} \\
	\And
	{\hspace{1mm}Andrei Filatov$^1$} \\
        \And
	{\hspace{1mm}Viacheslav Vasilev$^1$} \\
        \And
	{\hspace{1mm}Anastasia Maltseva$^1$} \\
        \And
        {\hspace{1mm}Said Azizov$^1$} \\
        \And
	{\hspace{1mm}Igor Pavlov$^1$} \\
        \And
        {\hspace{1mm}Julia Agafonova$^1$} \\
        \And
	{\hspace{1mm}Andrey Kuznetsov$^{1,2,*}$} \\
        \And
	{\hspace{1mm}Denis Dimitrov$^{1,2,*}$} \\
}

\renewcommand{\shorttitle}{Kandinsky 3.0 Technical Report}

\begin{document}
\maketitle

\begin{abstract}

{We}\blfootnote{$^*$Corresponding authors: Andrey Kuznetsov \href{kuznetsov@airi.net}{$<$kuznetsov@airi.net$>$}, Denis Dimitrov \href{dimitrov@airi.net}{$<$dimitrov@airi.net$>$}.} present {Kandinsky 3.0}\blfootnote{$^{**}$The family of models is named after Wassily Kandinsky, the great Russian artist and an art theorist, the father of abstract art.}, a large-scale text-to-image generation model based on latent diffusion, continuing the series of text-to-image Kandinsky models and reflecting our progress to achieve higher quality and realism of image generation. 
In this report we describe the architecture of the model, the data collection procedure, the training technique, and the production system for user interaction. We focus on the key components that, as we have identified as a result of a large number of experiments, had the most significant impact on improving the quality of our model compared to the others.
We also describe extensions and applications of our model, including super resolution, inpainting, image editing, image-to-video generation, and a distilled version of Kandinsky 3.0 -- Kandinsky 3.1, which does inference in 4 steps of the reverse process and 20 times faster without visual quality decrease.
By side-by-side human preferences comparison, Kandinsky becomes better in text understanding and works better on specific domains. The code is available at \url{https://github.com/ai-forever/Kandinsky-3}.

\end{abstract}

\section{Introduction}

Recently, the quality of text-to-image generation models increased significantly, which became possible thanks to the invention of Diffusion Probabilistic Models \cite{Dickstein2015, ho2020denoising}. To date, the zoo of text-to-image generation models is extremely rich \cite{Nichol2022glide, ramesh2022hierarchical, rombach2022high, saharia2022photorealistic, balaji2022eDiff, Midjourney, podell2023sdxl, dalle3}. Some of these systems provide users with opportunities for almost real-time inference, a high level of photorealism, an understanding of fantastic ideas and concepts not found in the real world, and many user-friendly web platforms and interfaces for a generation. Despite this, the task of text-to-image generating continues to pose serious challenges to researchers. The growing number of practical applications in commerce and design leads to a new unprecedented level of realism and alignment with complex textual descriptions.

This paper presents Kandinsky 3.0, a new text-to-image generation model based on latent diffusion \cite{rombach2022high}. Earlier, we introduced  other models of the Kandinsky family \cite{razzhigaev2023kandinsky}, the architecture of which is based on a two-stage pipeline using Diffusion Mapping between elements of latent vector spaces of images and text with subsequent decoding. In the Kandinsky 3.0 model, we focused on improving the text understanding, the image quality and simplifying the architecture by providing a single-stage pipeline in which generation takes place directly using text embeddings without any additional priors. The whole pipeline contains 11.9 billion parameters,  almost three times more than the largest of the previous models of the Kandinsky family. Also, we integrated Kandinsky 3.0 into our user-friendly website interaction system. We made our model completely public to promote the development of new technologies and openness in the scientific community.

This technical report is arranged as follows:

\begin{itemize}
    \item Firstly, we describe the user interaction system (Section \ref{sec:demo_system});
    \item Secondly, we describe the key components of the Kandinsky 3.0 model architecture (Section \ref{sec:Architecture}), dataset usage strategy and training techniques (Section \ref{sec:data_and_training});
    \item We also report additional features such as distillation (aka Kandinsky 3.1, see Section \ref{sec:Distilled}), super resolution (aka Kandinsky SuperRes, Section \ref{sec:SuperRes}) and prompt beautification (Section \ref{sec:beautification}), and various applications: inpainting (Section \ref{sec:Inpainting}), image editing (Section \ref{sec:editing}), image-to-video (Section \ref{sec:I2V}) and text-to-video (Section \ref{sec:T2V});
    \item Finally, we describe human evaluation methodology (Section \ref{sec:HumanEval}) and report the extensive results of side-by-side comparisons based on human preferences (see in appendix \nameref{sec:SBS}).
\end{itemize}

\section{Interaction System}\label{sec:demo_system}

As in the previous work \cite{razzhigaev2023kandinsky}, we incorporated the Kandinsky 3.0 model in several user interaction systems with open free access. Here we will describe their functionality and capabilities.

Fusionbrain.ai\footnote{\url{https://fusionbrain.ai/en/}} -- this is a web-editor that has the following functionality for text-to-image generation\footnote{A detailed description of the API can be found at \url{https://fusionbrain.ai/docs/en/doc/api-dokumentaciya/}.}:
\begin{itemize}
    \item The system can accept text prompts in Russian, English and other languages. It is also allowed to use emoji in the text description. The maximum size of a text is 1000 characters;
    \item In the ``Negative prompt'' field, the user can specify which information (e.g., colors) the model should not use for generation;
    \item Maximum resolution is $1024 \times 1024$;
    \item Choosing the sides ratio: $1:1$, $16:9$, $9:16$, $2:3$ or $3:2$;
    \item Choosing of generation style to accelerate inference: digital image, pixel art, cartoon, portrait photo, studio photo, cyberpunk, 3D render, classicism, anime, oil painting, pencil drawing, Khokhloma painting style, and styles of famous artists such as Aivazovsky, Kandinsky, Malevich, and Picasso;
    \item Zoom in/out;
    \item Using an eraser to highlight areas that can be filled both with and without a new text description (inpainting technique);
    \item Using a sliding window to expand the boundaries of the generated image and further generation with new borders (outpainting approach);
    \item We also implemented a content filter developed by us to process incorrect requests.
\end{itemize}

This website also supports image-to-video generation with the following characteristics:
\begin{itemize}
    \item Resolution: $640\times 640$, $480\times 854$ and $854\times 480$;
    \item The user can set up to 4 scenes by describing each scene using a text prompt. Each scene lasts 4 seconds, including the transition to the next;
    \item For each scene, it is possible to choose the direction of camera movement: up, down, right, left, counterclockwise or clockwise, zoom-in, zoom-out, and different types of flights around the object;
    \item The average generation time ranges from 1.5 minutes for one scene to 6 minutes for four scenes;
    \item The generated video can be downloaded in mp4 format.
\end{itemize}

We also created a Telegram-bot\footnote{\url{https://t.me/kandinsky21_bot}} in which text-to-image generation is available.

\section{Kandinsky 3.0 Architecture}\label{sec:Architecture}

\subsection{Overall pipeline}

\begin{figure}[ht]
    \center{\includegraphics[bb=0 0 2309 549, scale=0.19]{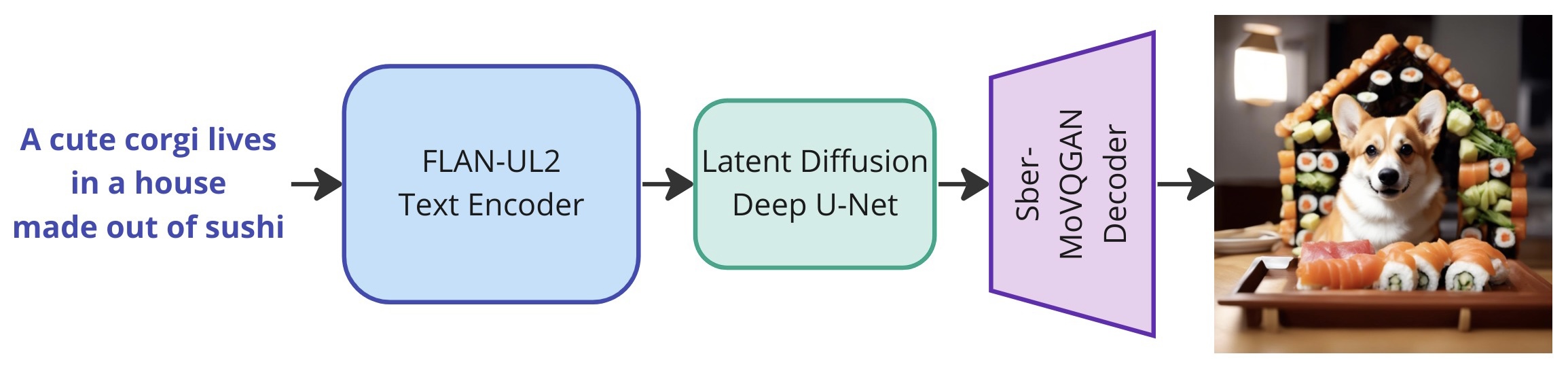}}
    \caption{\textbf{Kandinsky 3.0 overall pipeline architecture.} It consists of a text encoder, a latent conditioned diffusion model, and an image decoder.}
    \label{fig:full_pipeline}
\end{figure}

Kandinsky 3.0 is a latent diffusion model, the whole pipeline of which includes a text encoder for processing a prompt from the user, a U-Net \cite{ronneberger2015u} for predicting noise during the denoising (reverse) process, and a decoder for image reconstruction from the generated latent (Fig. \ref{fig:full_pipeline}). The text encoder and image decoder were frozen during the U-Net training. The whole model contains 11.9 billion parameters. Table \ref{tab:parameters} shows the number of parameters for the components of the Kandinsky 3.0 model, the Kandinsky 2.2 \cite{razzhigaev2023kandinsky} and SDXL \cite{podell2023sdxl} models. Below, we take a look at each component of our new model.

\begin{table}[H]
\centering
\small
\begin{tabular}{llll}
    \toprule
     & Kandinsky 2.2 \cite{razzhigaev2023kandinsky} & SDXL \cite{podell2023sdxl} & Kandinsky 3.0\\
    \midrule
    Model type  & Latent Diffusion & Latent Diffusion & Latent Diffusion \\
    Total parameters & 4.6B & 3.33B & 11.9B \\
    Text encoder & 0.62B & 0.8B & 8.6B \\
    Diffusion Mapping \cite{razzhigaev2023kandinsky} & 1.0B & -- & -- \\
    Denoising U-Net & 1.2B & 2.5B & 3.0B \\
    Image decoder & 0.08B & 0.08B & 0.27B \\
    \bottomrule
\end{tabular}
\caption{Comparison of the number of parameters (in billions) of components for the Kandinsky 2.2, SDXL, and Kandinsky 3.0 models.}   
\label{tab:parameters}
\end{table}

\subsection{U-Net architecture}\label{sec:unet}

This section, we describe the considerations and ideas that led us to create our denoising U-Net architecture. Based on the success of large transformer-based models in vision problems when learning on large amounts of data \cite{dosovitskiy2021an, Liu2021swin, Radford2021, ramesh2021dalle}, and the fact that convolutional architectures occupied the central place in diffusion models so far, we had to decide which types of layers would contain the main part of the parameters of the new model: transformer or convolutional. Our field of view mainly included classification model architectures that showed high quality on the ImageNet benchmark dataset \cite{Deng2009ImageNet}. We conducted about half a thousand experiments with various architectural combinations and noted the following two key points:

\begin{itemize}
    \item Increasing the network depth by increasing the number of layers while reducing the total number of parameters in practice gives better results in training. A similar idea of residual blocks with bottlenecks was previously exploited in the ResNet-50 \cite{He2016resnet} and BigGAN-deep architecture \cite{brock2019large};
    \item At the first network layers, in high resolution level, we decided to process the latents using only convolutional blocks, while more compressed latent representations were fed to the transformer layers too. This ensures the global interaction of image elements.
\end{itemize}

We also reviewed the MaxViT architecture \cite{tu2022maxvit}, which is almost entirely based on transformer blocks adapted to work with images by reducing the quadratic complexity of self-attention. In the classification task, this architecture shows the best results in comparison with the models mentioned above. Despite this, during experiments, we found out that this architecture does not show good results in the generation task.

\begin{figure}[ht]
    \center{\includegraphics[bb=0 0 1987 1050, scale=0.23]{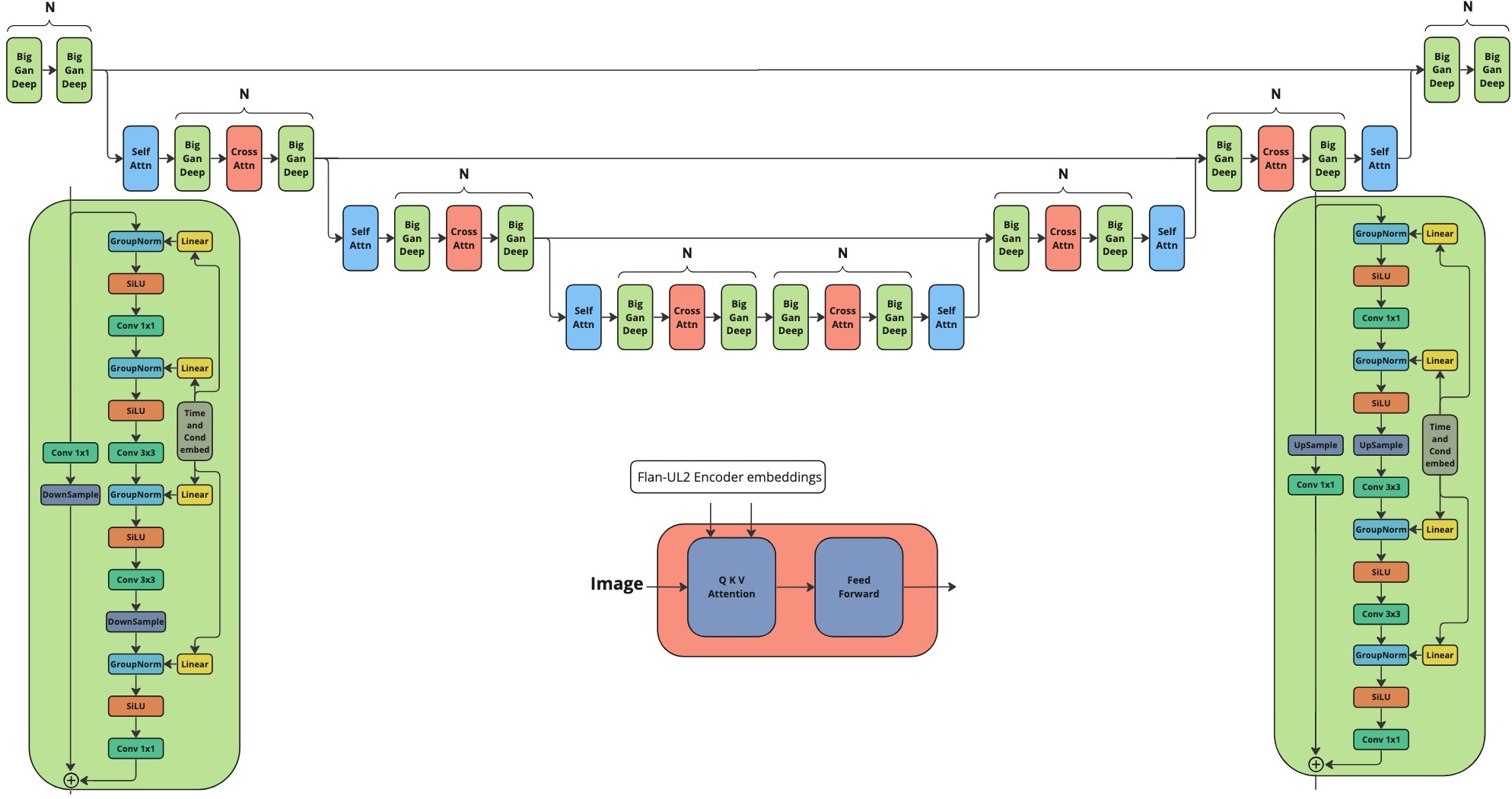}}
    \caption{\textbf{Kandinsky 3.0 U-Net architecture.} The architecture is based on modified BigGAN-deep blocks (left and right -- downsample and upsample versions), which allows us to increase the depth of the architecture due to the presence of bottlenecks. The attention layers are arranged at levels with a lower resolution than the original image.}
    \label{fig:UNet}
\end{figure}

Having thus explored all of the above architectures, we settled on the ResNet-50 block as the main block for our denoising U-Net. Thus, the residual blocks of our architecture at the input and output contain convolutional layers with a $1\times 1$ kernel, which correspondingly reduce and increase the number of channels. We also expanded it with one more convolutional layer with a $3\times 3$ kernel, as in the BigGAN-deep residual block architecture. Using bottlenecks in residual blocks made it possible to double the number of convolutional layers, while maintaining approximately the same number of parameters as without bottlenecks. At the same time, the depth of our new architecture has increased by 1.5 times compared to previous versions of the Kandinsky 2.X model.

At the higher levels of the upscale and downsample parts, we placed only our implementation of convolutional residual BigGAN-deep blocks. At the same time, at lower resolutions, the architecture includes self-attention and cross-attention layers. The complete scheme of our U-Net architecture, residual BigGAN-deep blocks, and cross-attention blocks is shown in Fig. \ref{fig:UNet}.

Our version of the BigGAN-deep residual blocks differs from the one proposed in \cite{brock2019large} by the following components:

\begin{itemize}
    \item We use Group Normalization \cite{GroupNorm2018} instead of Batch Normalization \cite{Ioffe2015batchNorm};
    \item We use SiLU \cite{elfwing2017sigmoidweighted} instead of ReLU \cite{agarap2019deep};
    \item As skip connections, we implement them in the standard BigGAN residual block. For example, in the upsample part of the U-Net, we do not drop channels but perform upsampling and apply a convolution with $1\times 1$ kernel.
\end{itemize}

\subsection{Text encoder}

For the text encoder, we use the 8.6B encoder of the Flan-UL2 20B model \cite{FlanUL2blogpost}, which
is based on the pre-trained UL2 20B \cite{Tay2022UL2UL}. In addition to pretraining on a large corpus of texts, Flan-UL2 was also trained using supervised fine-tuning on many language tasks using Flan Prompting \cite{chung2022scaling}. Our experiments showed that such fine-tuning also significantly improves image generation.

\subsection{Sber-MoVQGAN}

To achieve a high-quality image reconstruction in complex domains such as text and faces, we developed the Sber-MoVQGAN autoencoder, which showed good results in Kandinsky 2.2 \cite{razzhigaev2023kandinsky}. 

The Sber-MoVQGAN architecture is based on the VQGAN \cite{esser2021taming} architecture with the addition of spatially conditional normalization from the MoVQ \cite{zheng2022movq}. Spatial conditional normalization is implemented similarly to the Adaptive Instance Normalization (AdaIN) layers used in the StyleGAN \cite{Karras2018} architecture and is calculated by the formula:
\begin{equation}
     F^i=\phi_\gamma(z_q)\frac{F^{i-1}-\mu(F^{i-1})}{\sigma(F^{i-1})} + \phi_\beta(z_q)
\end{equation}
where $z_q$ is the latent consisting of codebook codes, $F^{i-1}$ is the intermediate feature map, $\mu$ and $\sigma$ are the functions for calculating the mean and standard deviation of the activation, $\phi_\gamma$ and $\phi_\beta$ are the trainable affine transformations, which convert $z_q$ into the scaling and bias values. 
Other important features of our implementation include the addition of EMA (exponential moving average) weights and a modified loss function from ViT-VQGAN \cite{yu2022vectorquantized} during the training stage.

We trained three versions of Sber-MoVQGAN -- 67M, 102M, and 270M. The 67M version is the same size as the standard VQGAN. The 102M model uses twice the number of residual blocks compared to the 67M, and the 270M model operates with twice the original number of channels. Kandinsky 3.0 uses the 270M model as the image decoder. 

We trained Sber-MoVQGAN on the LAION HighRes dataset \cite{Schuhmann2022}, obtaining the SOTA results in image reconstruction. The comparison of our autoencoder with competitors and Sber-VQGAN\footnote{\url{https://github.com/ai-forever/tuned-vq-gan}} are presented in Table \ref{tab:model_comparisons}. We released the weights and code for these models under an open-source license \footnote{\url{https://github.com/ai-forever/MoVQGAN}}.

\begin{table*}[hbt!]
\small
\centering
\caption{Sber-MoVQGAN comparison with competitors on ImageNet dataset. Our 270B version outperform all other models in terms of all quality metrics.}
\label{tab:model_comparisons}
\begin{tabular}{llllllll}
\hline
\bf Model & \bf  Latent size & \bf  Num Z & \bf  Train steps & \bf  FID $\downarrow$ & \bf  SSIM $\uparrow$ & \bf  PSNR $\uparrow$ & \bf  L1 $\downarrow$ \\
\hline
ViT-VQGAN \cite{yu2022vectorquantized} & 32x32 & 8192 & 500,000 & 1.28 & -- & -- & -- \\
RQ-VAE \cite{lee2022autoregressive} & 8x8x16 & 16384 &  10 epochs & 1.83 & -- & -- & -- \\
Mo-VQGAN  \cite{zheng2022movq} & 16x16x4 & 1024 &  40 epochs & 1.12 & 0.673 & 22.42 & -- \\
VQ CompVis \cite{Blattmann2022compvis} & 32x32 & 16384 & 971,043 & 1.34 & 0.650 & 23.85 & 0.0533 \\
KL CompVis \cite{Blattmann2022compvis} & 32x32 & -- & 246,803 & 0.968 & 0.692 & 25.11 & 0.0474 \\
Sber-VQGAN & 32x32 & 8192 & 1 epoch & 1.44 & 0.682 & 24.31 & 0.0503 \\
Sber-MoVQGAN 67M & 32x32 & 1024 & 5,000,000 & 1.34 & 0.704 & 25.68 & 0.0451 \\
Sber-MoVQGAN 67M & 32x32 & 16384 & 2,000,000 & 0.965 & 0.725 & 26.45 & 0.0415 \\
Sber-MoVQGAN 102M & 32x32 & 16384 & 2,360,000 & 0.776 & 0.737 & 26.89 & 0.0398 \\
Sber-MoVQGAN 270M & 32x32 & 16384 & 1,330,000 & \textbf{0.686} & \textbf{0.741} & \textbf{27.04} & \textbf{0.0393}\\\hline
\label{movq}
\end{tabular}
\end{table*}

\section{Data and Training Strategy}\label{sec:data_and_training}

\paragraph{Data.} During the training procedure, we used a large dataset of text-image pairs collected online. The training dataset consists of popular open-source datasets and our internal dataset of approximately 150 million text-image pairs. To improve data quality, we pass the data through several filters: the aesthetics quality of the image, the watermarks detection, the CLIP similarity of the image with the text \cite{Radford2021}, and the detection of duplicates with perceptual hash.

We discovered that the collected data from Common Crawl \cite{commoncrawl} contains almost no images related to Russian culture. To fix this, we collected and labeled a dataset of 200 thousand text-image pairs from Soviet and Russian cartoons, famous people, and places. This dataset helped improve the model's quality and text alignment when generating Russian-related images.

We also divided all the data into two categories. We used the first at the initial stages of low-resolution pretraining and the second for mixed and high-resolution fine-tuning at the last stage. The first category includes open large text-image datasets such as LAION-5B \cite{schuhmann2022laion5b} and COYO-700M \cite{kakaobrain2022coyo} and ``dirty'' data that we collected from the Internet. The second category contains the same datasets but with stricter filters, especially for the image aesthetics quality.

\paragraph{Training.} We divided the training process into several stages to use more data and train the model to generate images in a wide range of resolutions:

\begin{enumerate}
    \item $\mathbf{256 \times 256}$ \textbf{resolution:} 1.1 billions of text-image pairs, batch size $= 20$, 600 thousand steps, 104 NVIDIA Tesla A100;
    \item $\mathbf{384 \times 384}$ \textbf{resolutions:} 768 millions of text-image pairs, batch size $= 10$, 500 thousand steps, 104 NVIDIA Tesla A100;
    \item $\mathbf{512 \times 512}$ \textbf{resolutions:} 450 millions of text-image pairs, batch size $= 10$, 400 thousand steps, 104 NVIDIA Tesla A100;
    \item $\mathbf{768 \times 768}$ \textbf{resolutions:} 224 millions of text-image pairs, batch size $= 4$, 250 thousand steps, 416 NVIDIA Tesla A100;
    \item \textbf{Mixed resolution:} $\mathbf{768^2 \leq W\times H \leq 1024^2}$, 280 millions of text-image pairs, batch size $= 1$, 350 thousand steps, 416 NVIDIA Tesla A100.
\end{enumerate}

\section{Additional Features}

\subsection{Kandinsky 3.1 (aka Kandinsky 3.0 Distilled)}\label{sec:Distilled}

\begin{figure}[ht]
    \center{\includegraphics[bb=0 0 1991 1285, scale=0.2]{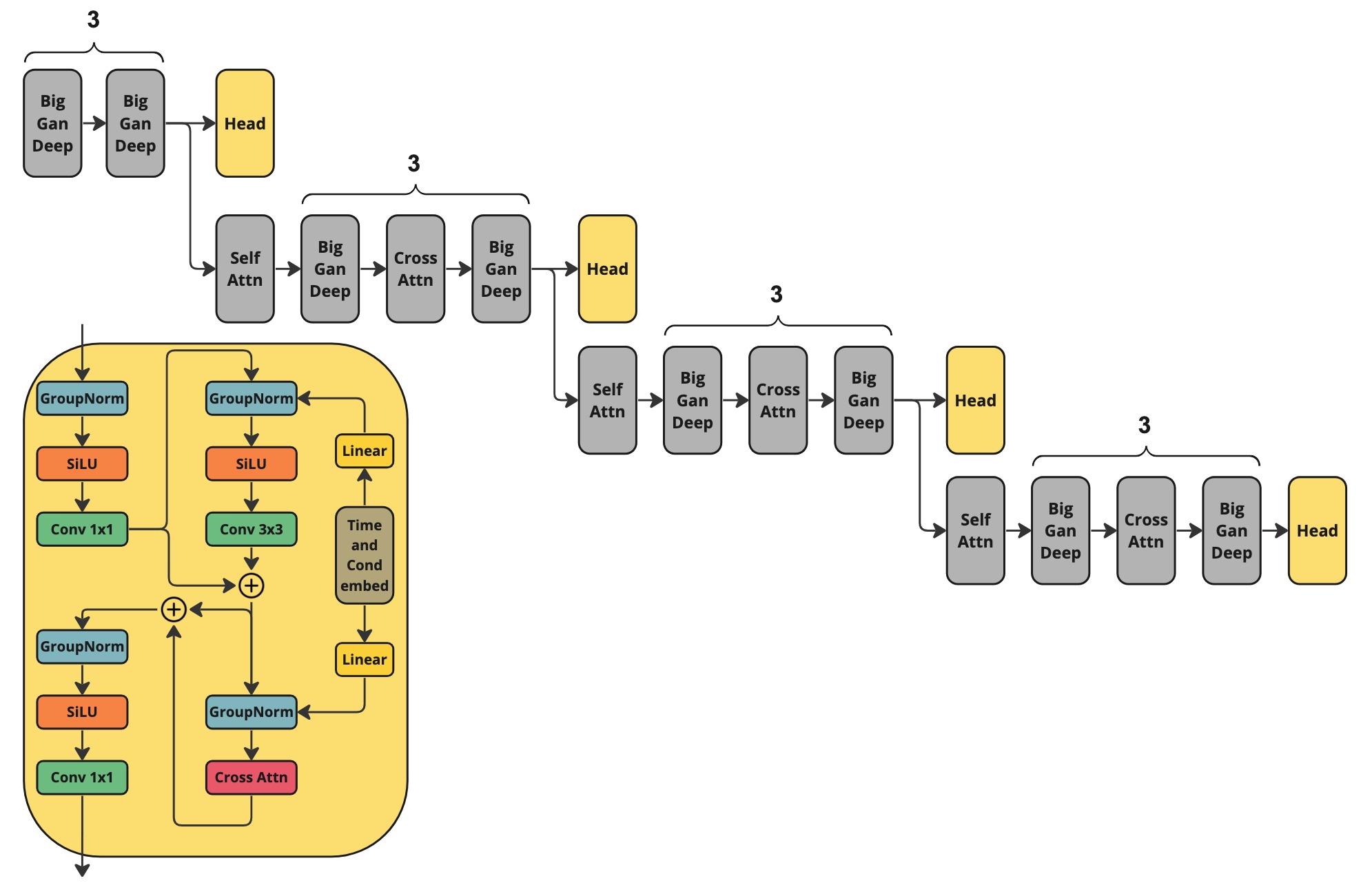}}
    \caption{\textbf{Discriminator architecture for Kandinsky 3.1.} Gray blocks inherit the weight of Kandinsky 3.0 and remain frozen during training.}
    \label{fig:discriminator}
\end{figure}

A serious problem of diffusion models is the generation speed. To obtain a single image, it is usually necessary to go through several dozen steps (for example, 50) in the reverse process, that is, to pass data through U-Net many times with a batch size of 2 (with conditioning and without it) for classifier free guidance. To solve this problem, we used the Adversarial Diffusion Distillation approach \cite{sauer2023adversarial}, but with a number of significant modifications:

\begin{enumerate}
    \item If pretrained pixel models were used as a discriminator, it would be necessary to decode the generated image using MoVQ and throw gradients through it, which would lead to huge memory costs. These costs do not allow training the model in $1024\times 1024$ resolution. \textbf{Therefore, as a discriminator, we used the frozen downsample part of the U-Net from Kandinsky 3.0} with trainable heads after each layer of resolution reduction (Fig. \ref{fig:discriminator});
    \item We added cross-attention on text embeddings from FLAN-UL2 to the discriminator heads instead of adding text CLIP-embeddings. \textbf{This improved the text alignment} using a distilled model;
    \item \textbf{We used Wasserstein Loss} \cite{pmlr-v70-arjovsky17a}. Unlike Hinge Loss, it is unsaturated, which avoids the problem of zeroing gradients at the first stages of training, when the discriminator is stronger than the generator;
    \item \textbf{We removed the regularization in the Distillation Loss}, since according to our experiments it did not affect the quality of the model;
    \item We found that the generator quickly becomes more powerful than the discriminator, which leads to learning instability. To solve this problem, \textbf{we have significantly increased the learning rate of the discriminator}. For the discriminator, we set the learning rate is equal to $1e-3$, and for the generator $1e-5$. To prevent divergence, we also used gradient penalty, as in the \cite{sauer2023adversarial}.
\end{enumerate}

We trained a distilled model on a dataset with 100 thousand of highly-aesthetic image-text pairs, which we manually selected from the pretraining dataset. \textbf{As a result, we speed up the Kandinsky 3.0 by almost 20 times, making it possible to generate an image in only 4 passes through U-Net.} The acceleration is also due to the fact that there is no need to use classifier free guidance in the distilled version. However, like in \cite{sauer2023adversarial}, for a serious acceleration, we had to sacrifice the quality of the text comprehension, which can be seen from the human evaluation results of side-by-side comparison (Section \nameref{sec:distillation-sbs-appendix}). Generation examples by Kandinsky 3.1 can be found in the sections \nameref{sec:Distillation-appendix} and \nameref{sec:comparison-examples-appendix}.

\subsection{Kandinsky SuperRes}\label{sec:SuperRes}

\begin{wrapfigure}{r}{0.5\textwidth}
     \center{\includegraphics[bb=0 0 2100 4200, scale=0.095]{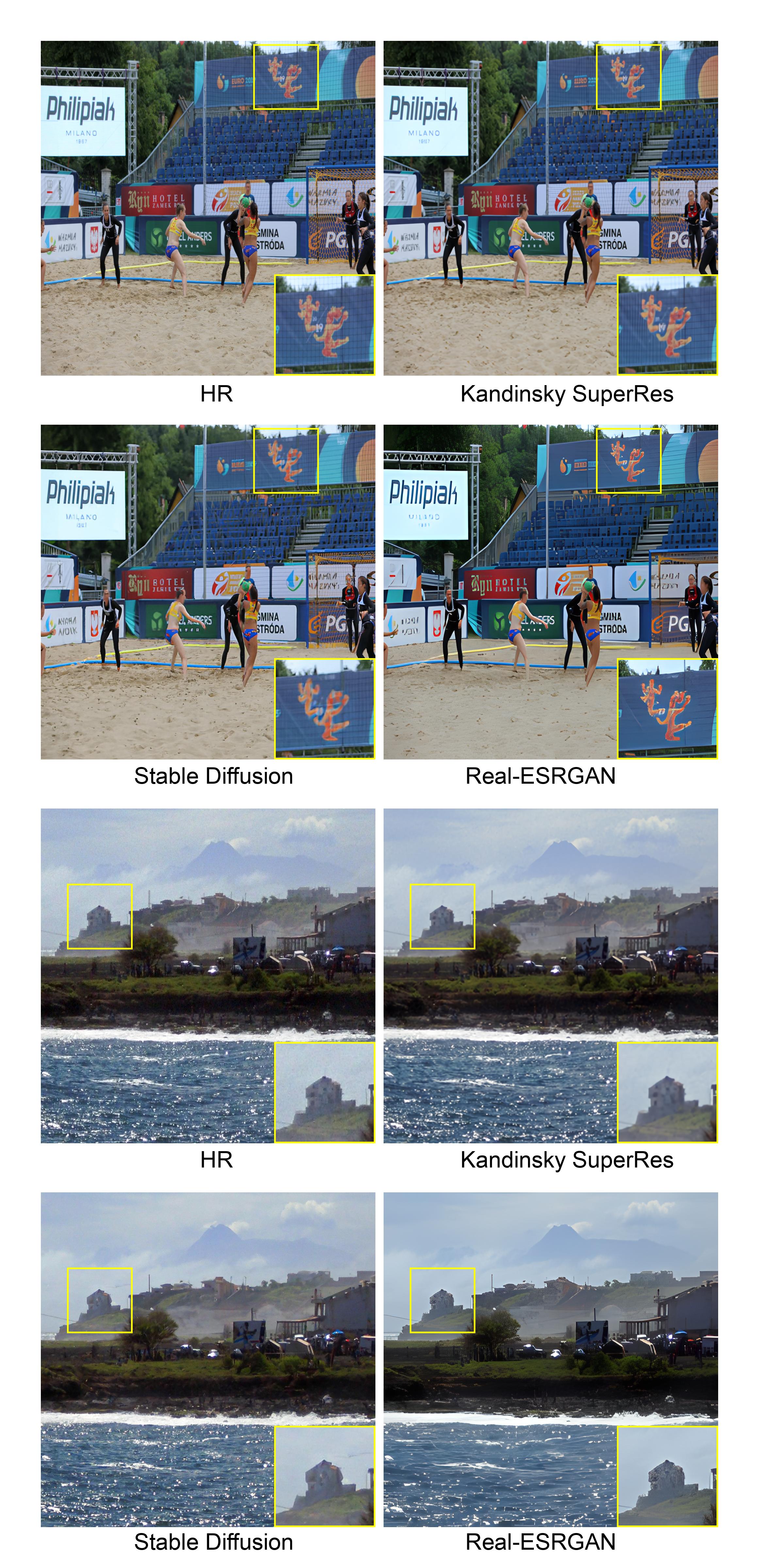}}
     \caption{\textbf{Comparison of Kandinsky SuperRes, Stable Diffusion and Real-ESRGAN models at 1024 resolution.} Better to zoom image.}
     \label{fig:comparison_sr}
\end{wrapfigure}

Based on Kandinsky 3.0, we developed the Kandinsky SuperRes model to generate high-resolution 4k images. Here we describe the modifications that we made.

\begin{enumerate}
    \item For Kandinsky SuperRes model we used diffusion in pixel space instead of latent diffusion. This eliminated the loss of quality when encoding and decoding images using an autoencoder. In addition, our experiments have shown that model based on pixel diffusion converges faster and better in this task than with latent diffusion.
    \item We implemented Efficient U-Net, similar to the one described in the Imagen \cite{saharia2022photorealistic}. Compared to U-Net from Kandinsky 3.0, Efficient U-Net consumes less memory and also has better convergence. Instead of 3 residual blocks at each downscaling, Efficient U-Net uses more low-resolution blocks and fewer high-resolution blocks. The order of convolution and downsampling/upsampling operations is changed relative to the original U-Net. In addition, we removed conditioning for text prompt, because it does not contribute to 4K high resolution generation. As a result, Efficient U-Net of Kandinsky SuperRes contains 413M parameters.
    \item During training, Efficient U-Net predict $x_0$ (i.e., the original image) instead of the noise level at a given time t, which avoided problems with changing the color of the generated SR image.
\end{enumerate}

The training was carried out in 2 stages. First, the model learned on the LAION dataset \cite{schuhmann2022laion5b} on 32 A100 for 1570 thousand steps with a batch size of 2 for a resolution of 256 to 1024. Then we trained the model on the aesthetic high-resolution sets, which we used to train Kandinsky 3.0 for 1500 thousand steps. At the second stage of training, we included JPEG compression similar to the scheme in the Real-ESRGAN \cite{wang2021realesrgan}.

The Kandinsky SuperRes model can work with images of various resolutions, but the main goal is 4K high-resolution generation. Since the Kandinsky SuperRes model was already trained for resolutions from 256 to 1024, higher resolution training was not possible due to GPU Tesla A100 memory overflow. For this reason, we used the MultiDiffusion algorithm \cite{bartal2023multidiffusion} to generate 4k images. More specifically, first we divide the image into overlapping patches, and then at each diffusion step we remove the noise and average the pixel/latent values of the overlapping areas. Thus, having gone through all the stages of diffusion, we obtain a seamless image of any resolution. Kandinsky SuperRes model works in inference in 5 steps using DPMSolverMultistepScheduler\footnote{\url{https://huggingface.co/docs/diffusers/api/schedulers/multistep_dpm_solver}}. The inference time for image generation in 4K resolution takes 13 seconds, and in 1K it takes 0.5 seconds.

\begin{figure}
     \center{\includegraphics[bb=0 0 4500 4500, scale=0.05]{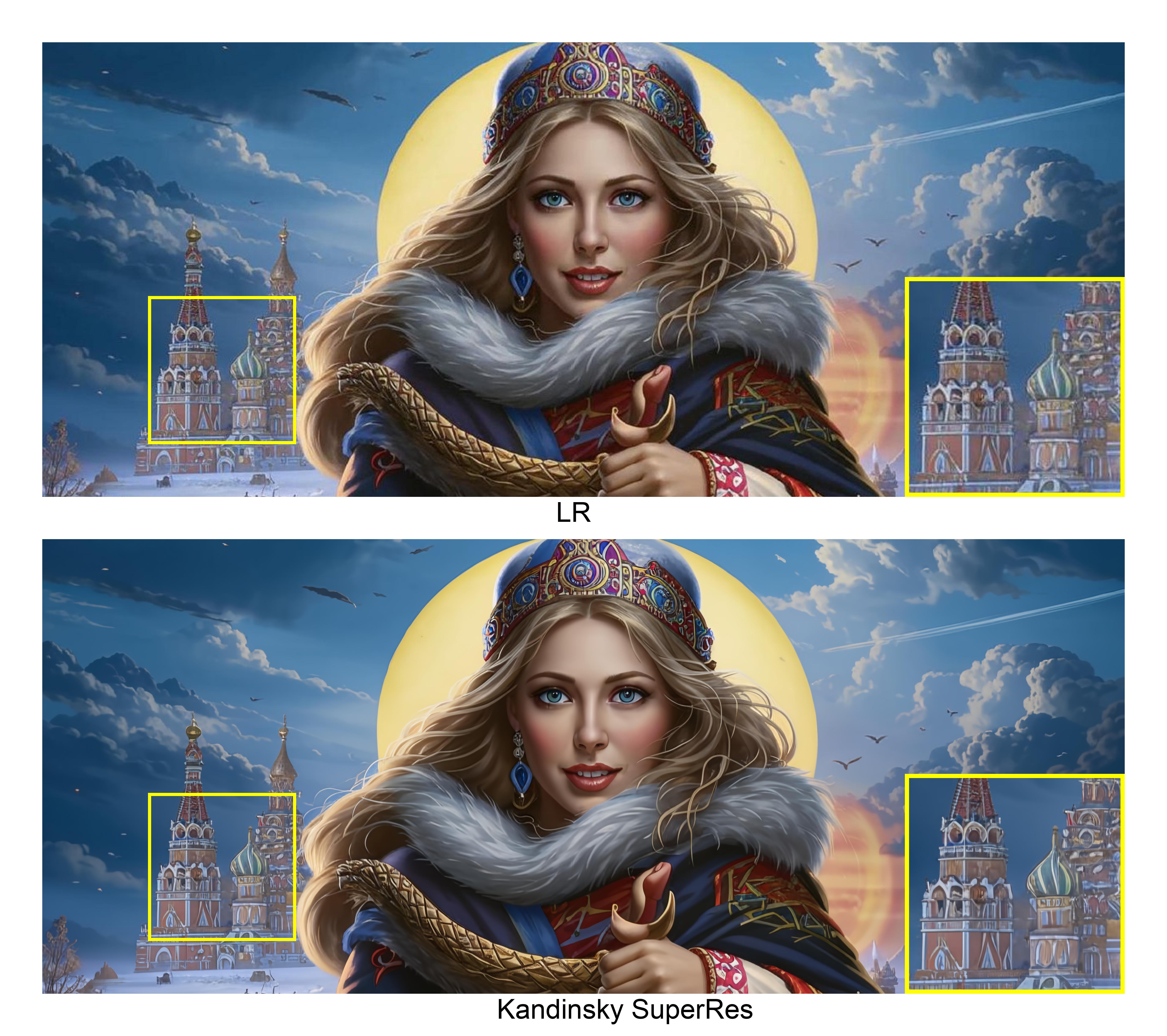}}
     \caption{\textbf{Example of Kandinsky SuperRes generation in 4K resolution.} Better to zoom image.}
     \label{fig:4K_sr}
\end{figure}

The table \ref{tab:kandisuperres} shows a comparison of Kandinsky SuperRes with the Real-ESRGAN \footnote{\url{https://github.com/ai-forever/Real-ESRGAN/}} and Stable Diffusion x4 Upscaler \footnote{\url{https://huggingface.co/stabilityai/stable-diffusion-x4-upscaler}} \cite{rombach2022high} models in terms of FID, SSIM, PSNR and L1 metrics on the our own dataset Wikidata 5K and RealSR(V3) \footnote{\url{https://github.com/csjcai/RealSR/tree/master}} \cite{cai2019toward} and Set14  \footnote{\url{https://paperswithcode.com/dataset/set14}} \cite{Huang-CVPR-2015} datasets. Our own dataset Wikidata 5K contains 5 thousand images collected from Wikipedia in 1K resolution. RealSR(V3) contains 100 test images in 1K and 2K resolutions. Set14 contains 14 low resolution images with JPEG artifacts. As a result, the Kandinsky SuperRes model shows the best results in terms of all quality metrics for all datasets. Figure \ref{fig:comparison_sr} shows examples of generation of Kandinsky SuperRes, Stable Diffusion and Real-ESRGAN models at a resolution of 1024. Figure \ref{fig:4K_sr} shows examples of generation of Kandinsky SuperRes in 4K resolution. We released the weights and code for Kandinsky SuperRes under an open-source license \footnote{\url{https://github.com/ai-forever/KandiSuperRes/}}.

\begin{table}
\centering
\caption{Comparison of Kandinsky SuperRes, Real-ESRGAN \cite{wang2021realesrgan} and Stable Diffusion \cite{rombach2022high} models. Kandinsky SuperRes outperforms other models in terms of all quality metrics for all test datasets.}
\label{tab:kandisuperres}
\begin{tabular}{llllll}
\hline
{Datasets} &
  {Model} &
  {FID↓} &
  {PSNR↑} &
  {SSIM↑} &
  {L1↓} \\
  \hline
{} &
  {Real-ESRGAN} &
  {9.96} &
  {24.48} &
  {0.73} &
  {0.0428} \\
{} &
  {Stable Diffusion} &
  {3.04} &
  {25.05} &
  {0.67} &
  {0.0435} \\
\multirow{-3}{*}{{Wikidata 5k}} &
  {Kandinsky SuperRes} &
  {\textbf{0.89}} &
  {\textbf{28.52}} &
  {\textbf{0.81}} &
  {\textbf{0.0257}} \\
  \hline
{} &
  {Real-ESRGAN} &
  {73.26} &
  {23.12} &
  {0.72} &
  {0.0610} \\
{} &
  {Stable Diffusion} &
  {47.79} &
  {24.85} &
  {0.67} &
  {0.0493} \\
\multirow{-3}{*}{{RealSR(V3)\cite{cai2019toward}}} &
  {Kandinsky SuperRes} &
  {\textbf{47.37}} &
  {\textbf{25.05}} &
  {\textbf{0.75}} &
  {\textbf{0.0462}} \\
  \hline
{} &
  {Real-ESRGAN} &
  {115.94} &
  {22.88} &
  {0.62} &
  {0.0561} \\
{} &
  {Stable Diffusion} &
  {76.32} &
  {23.60} &
  {0.57} &
  {0.0520} \\
\multirow{-3}{*}{{Set14 \cite{Huang-CVPR-2015}}} &
  {Kandinsky SuperRes} &
  {\textbf{61.00}} &
  {\textbf{25.70}} &
  {\textbf{0.70}} &
  {\textbf{0.0390}} \\\hline
\end{tabular}
\end{table}

\subsection{Prompt beautification}\label{sec:beautification}

Many diffusion text-to-image models have some kind of inconvenience due to the fact that the visual quality and detail of the generation depends on the degree of detail of the text prompt. Sometimes, in practice, user has to use long, redundant prompts to generate desirable images. Normal user prompts are not such. To solve this problem, we have built a function into the generation pipeline to add details to the user's prompt using LLM. An instruction is sent to the input of the language model with a request to improve the prompt, and the model's response is sent as the input into Kandinsky 3.0 model. As an LLM, we used Neural-Chat-7b-v3-1 \cite{NeuralChatPost} (based on Mistral 7B \cite{jiang2023mistral}) with the following system instruction:

\texttt{\#\#\# System:\slashn{n}You are a prompt engineer. Your mission is to expand prompts written by user. You should provide the best prompt for text to image generation in English. \slashn{n}\#\#\# User:\slashn{n}\{prompt\}\slashn{n}\#\#\# Assistant:\slashn{n}}

Here \texttt{\{prompt\}} is the user's text prompt. Examples of generation for the same prompt with and without beautification are presented in the section \nameref{sec:PromptBeautification}. We conducted a side-by-side human evaluation comparison of the generation quality using the prompt beautification and without it. We conducted comparison for both Kandinsky 3.0 and Kandinsky 3.1 to assess how strongly the language model affects the generated images. Each generation was evaluated by visual quality and by the text-image alignment. The results of the comparison can be seen in the section \nameref{sec:prompt-beautification-sbs-appendix}. In general, human preferences are definitely more inclined towards generations with prompt beautification. The only exception is the correspondence of the generated image to the text for the Kandinsky 3.1 model. This is due to the fact that the distilled model generally understands the text worse.

\section{Applications}

\subsection{Inpainting and Outpainting}\label{sec:Inpainting}

Implementation of the inpainting model is the same as GLIDE \cite{nichol2021glide}: we initialize our model from base Kandinsky model weights. Then, we modify the input convolution layer of U-Net (Section \ref{sec:unet}) so that the input can additionally accept the image latent and mask. Thus, U-Net takes as many as 9 channels as input: 4 for the original latent, 4 for the image latent, and an additional channel for the mask. We zeroed the additional weights, so training starts with the base model.

\begin{figure}
            \centering
                \begin{tabular}{ccc}
                    \includegraphics[bb=0 0 1029 682, width=0.3\linewidth]{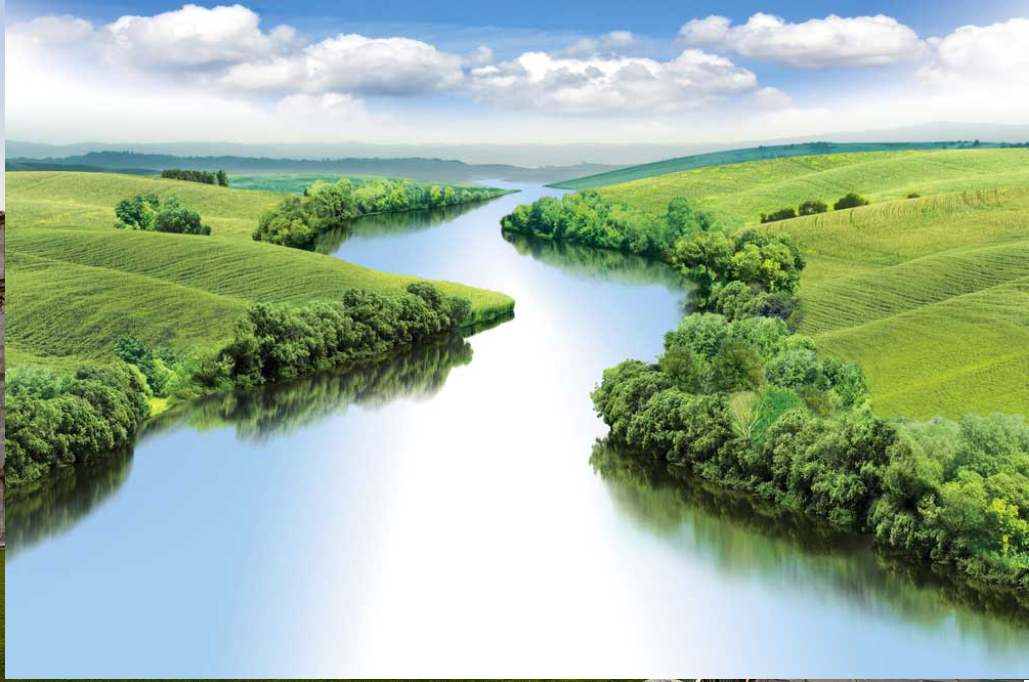}
                    & 
                    \hspace{4pt}\includegraphics[bb=0 0 1050 727, width=0.309\linewidth]{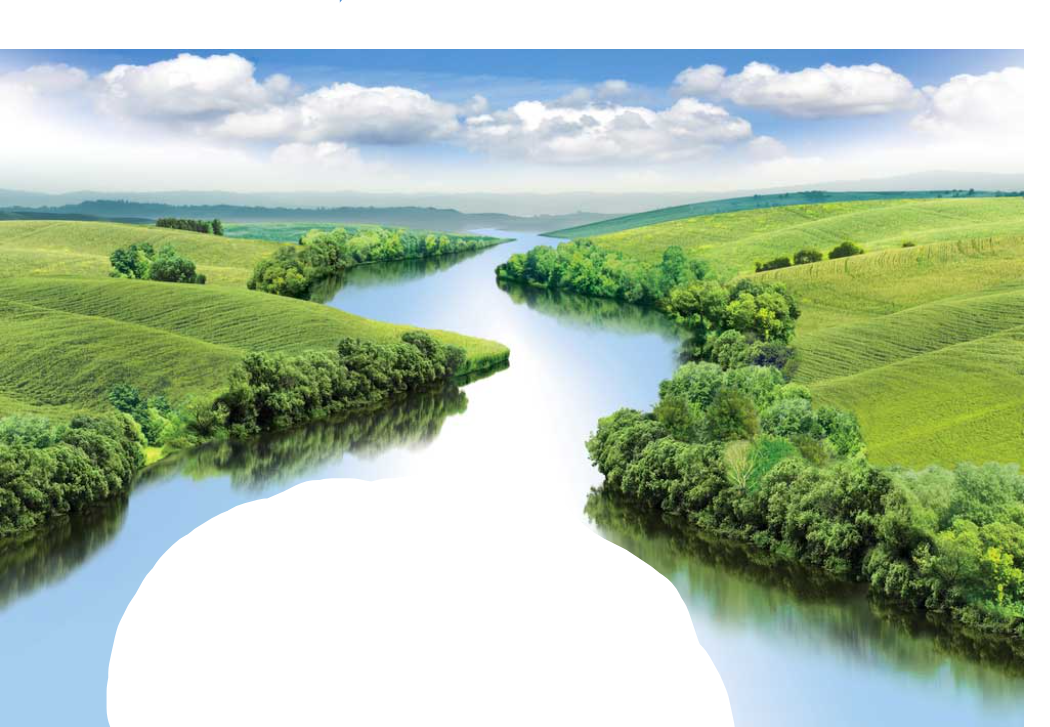}
                    
                    & 
                    \hspace{1pt}
                    \includegraphics[bb=0 0 1050 728, width=0.307\linewidth]{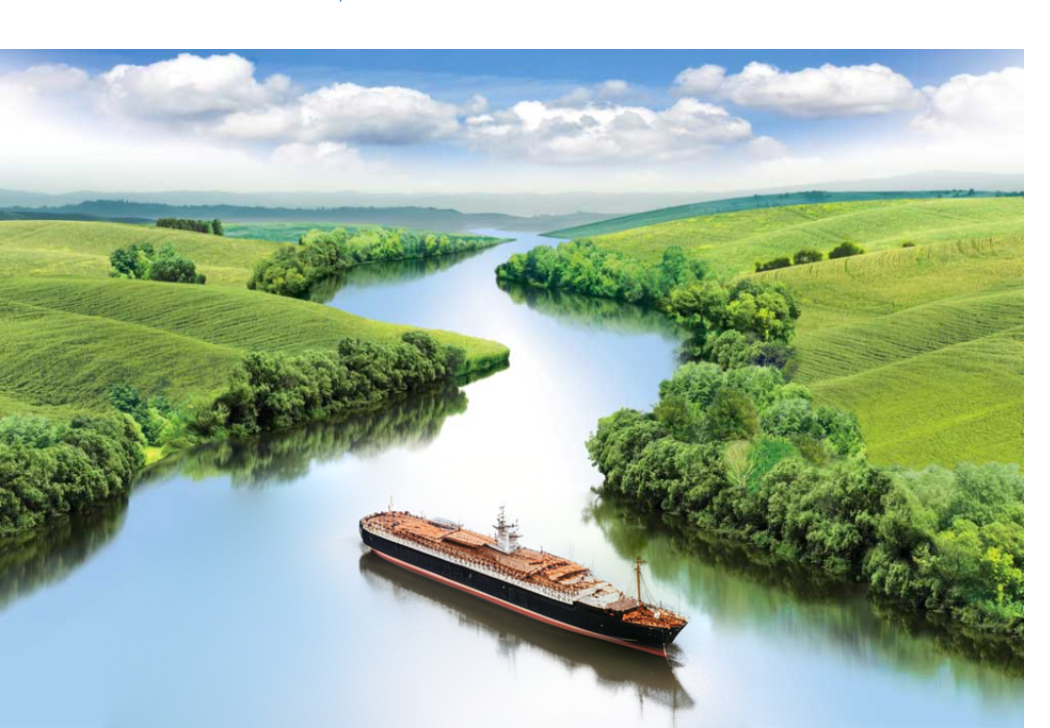}\\[-4pt]
                    \includegraphics[bb=0 0 1024 682, width=0.3\linewidth]{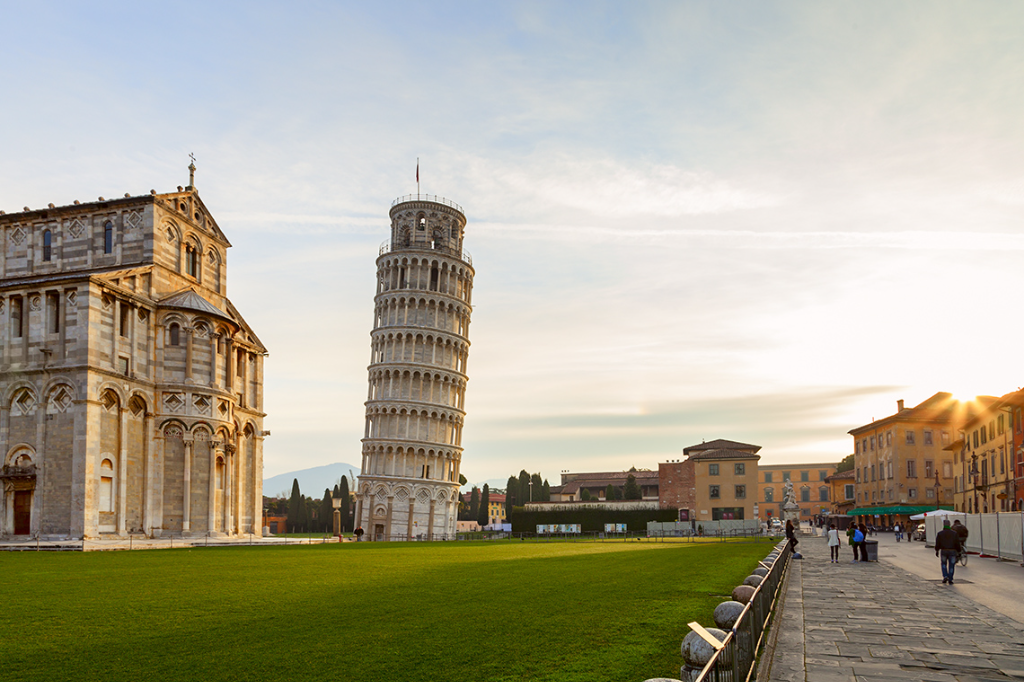}
                    & \includegraphics[bb=0 0 1024 682, width=0.3\linewidth]{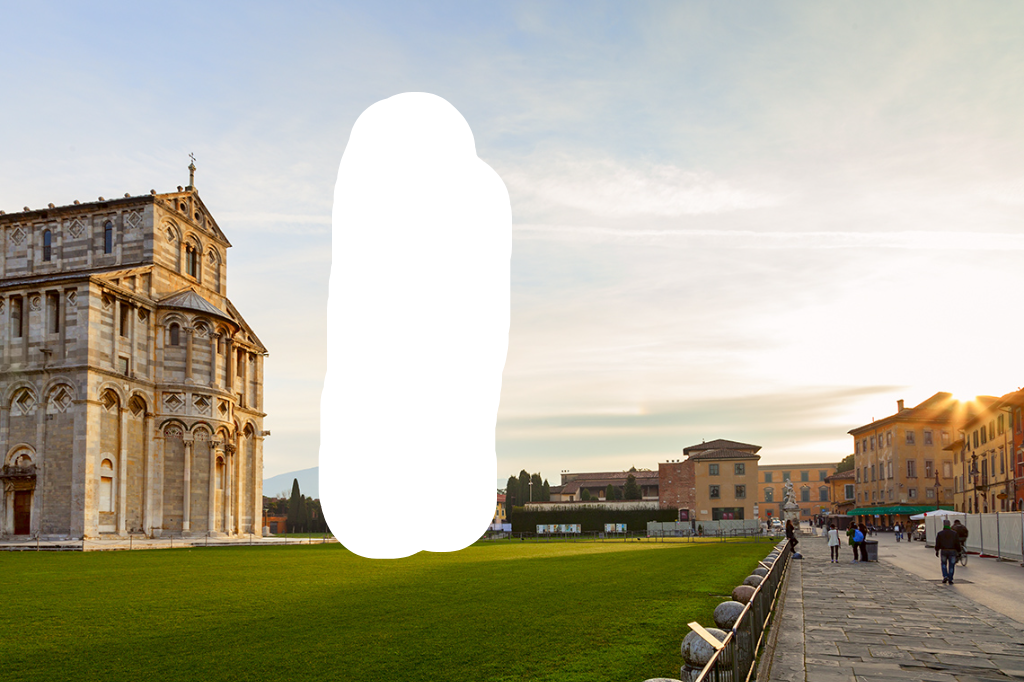}
                    & \includegraphics[bb=0 0 1029 682, width=0.3\linewidth]{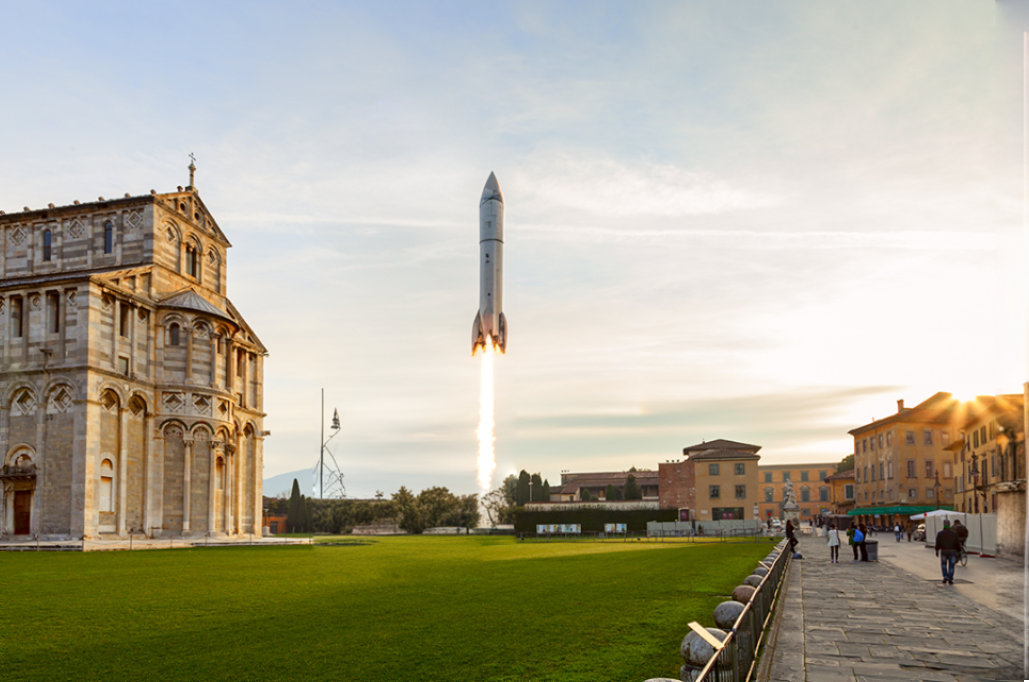}\\[-4pt]
                    \includegraphics[bb=0 0 1024 1024, width=0.3\linewidth]{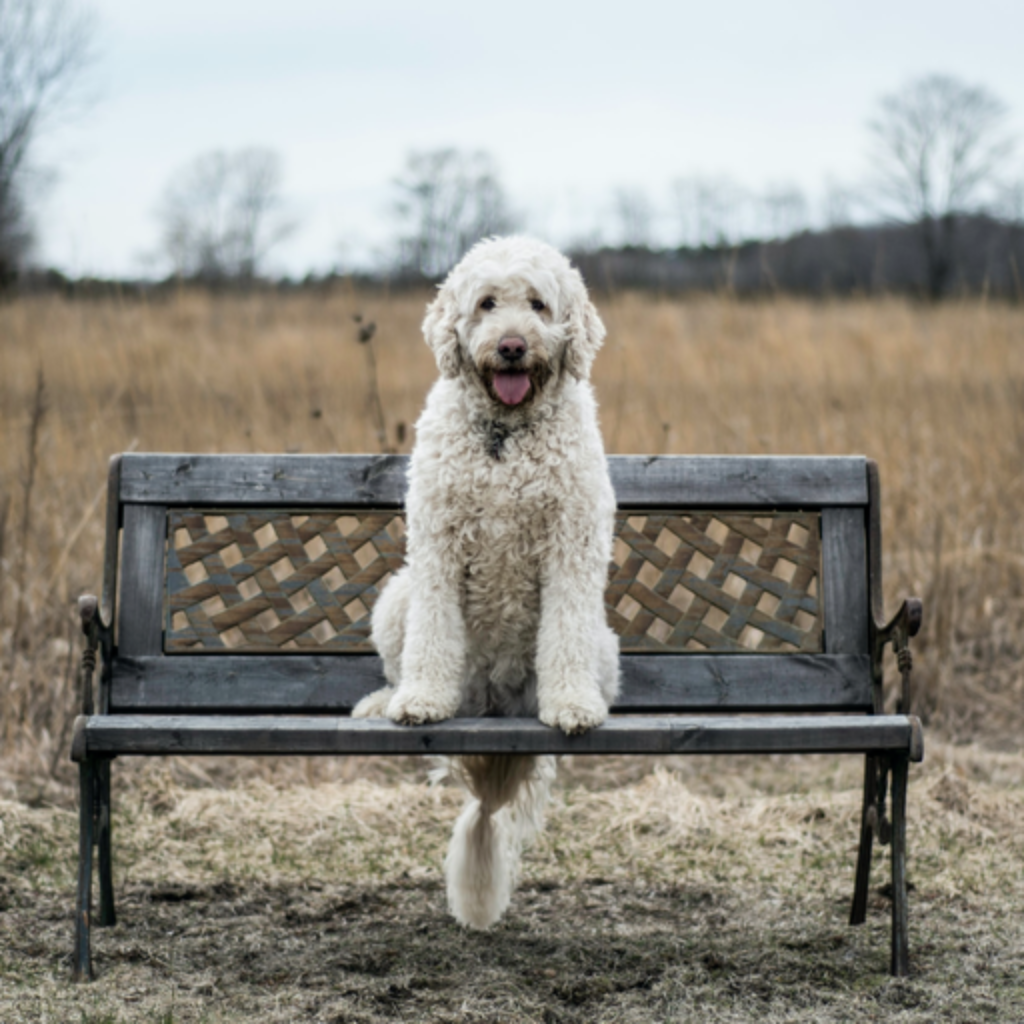}
                    & \includegraphics[bb=0 0 767 767, width=0.3\linewidth]{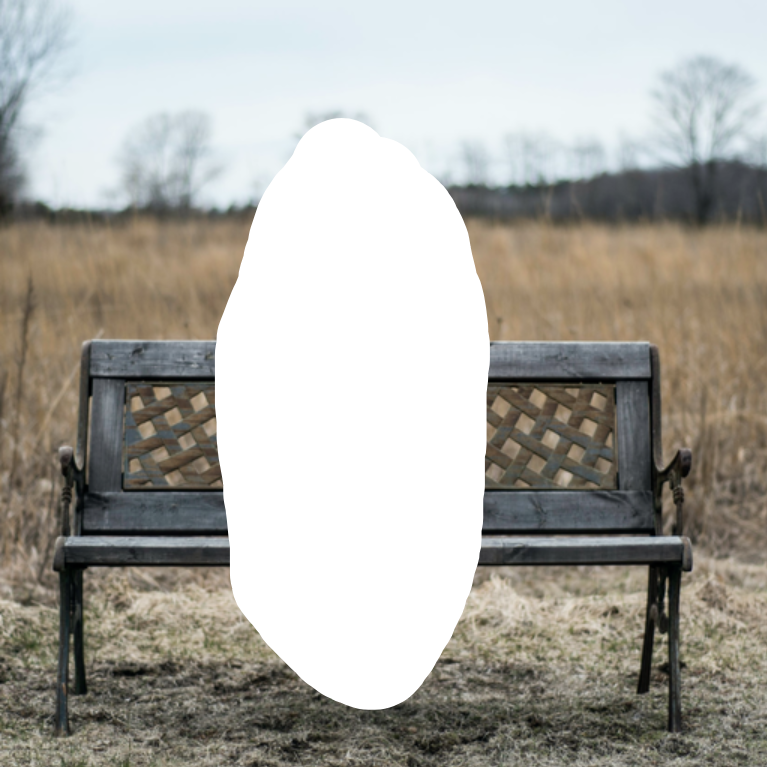}
                    & \includegraphics[bb=0 0 1024 1024, width=0.3\linewidth]{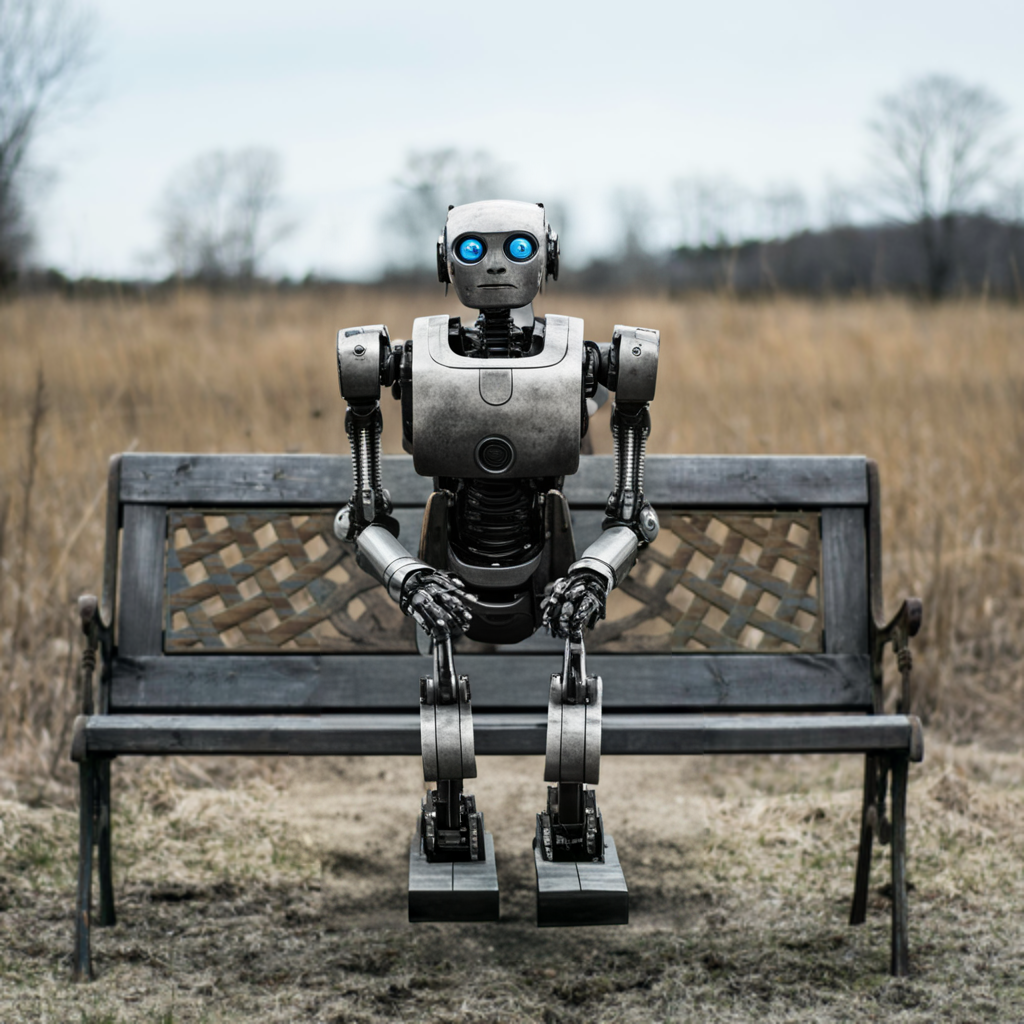}\\[-4pt]
                \end{tabular}%
             \caption{\textbf{Examples of inpainting.}\label{tab:picturegrid}. We use the following prompts for inpainting: \texttt{``ship sailing in the water''}, \texttt{``rocket''} and \texttt{``robot sitting on a bench''}, respectively. We generated inpainted images using  \protect\href{https://www.fusionbrain.ai}{FusionBrain.ai} platform.}
        \end{figure}

For training, we generate random masks of the following forms: rectangular, circles, strokes, and arbitrary form. For every image sample, we use up to 3 masks, and for every image, we use unique masks. We use the same dataset as for the training base model with generated masks. We train our model using Lion \cite{chen2023symbolic}
with lr=1e-5 and apply linear warmup for the first 10k steps. We train our model for 250 thousand steps. The inpainting results are in Fig. \ref{tab:picturegrid}. The outpainting results can be found in the \nameref{sec:outpainting_appendix} appendix section. Additionally, we finetune our model using object detection datasets and LLaVA captions for 50k steps.

\subsection{Image editing}\label{sec:editing}

Kandinsky 3.0 provides image generation not only using a text prompt, but also using an image as a visual prompt. This allows one to edit an existing image, change its style and add new objects to it. To do this, we extended an IP Adapter-based approach \cite{ye2023ip-adapter}. To implement our own IP-Adapter based on our basic generation model, we used attention adapters. We used the ViT-L-14 as the CLIP model \cite{Radford2021}. We get CLIP-embeddings of the size \texttt{batch size $\times$ 768}, which are then transformed by a linear layer into tensors of the size \texttt{batch size $\times$ 4 $\times$ 4096}. By adding a couple of new layers for key and value images in the cross-attention mechanism, we sum up the output of a text cross-attention with the output of cross-attention for images. We trained the IP-Adapter on the COYO 700m dataset \cite{kakaobrain2022coyo} with a batch size 288 during 800 thousand iterations. As a result, Kandinsky 3.0 supports image variation, image-image mixing, and image-text mixing. For generation examples see section \nameref{sec:IP-Adapter-appendix}.

In addition, we found that the IP Adapter-based approach does not preserve the shape of objects in the image, so we decided to train ControlNet \cite{zhang2023adding} in addition to our generation model to consistently change the appearance of the image, preserving more information compared to the original one. We used HED detector \cite{xie15hed} as a model to obtain the edges in the image fed to the ControlNet input. The training lasted 5 thousand iterations on the COYO 700m dataset \cite{kakaobrain2022coyo} on 8 Tesla A 100 GPU with a batch size 512.

\subsection{Image-to-Video Generation}\label{sec:I2V}

\begin{figure}[t]
  \centering
  \includegraphics[bb=0 0 800 592, scale=0.4]{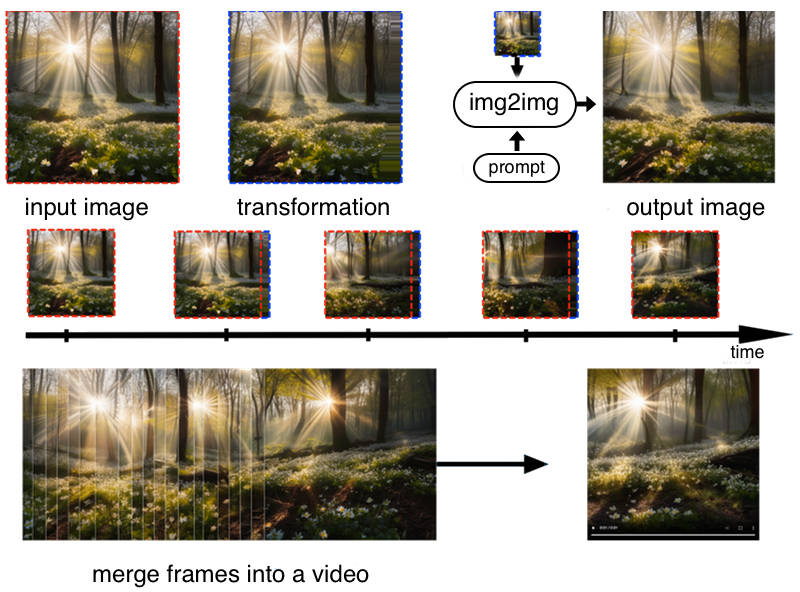}
  \caption{
  \textbf{Illustration of the Image-to-Video Generation process utilizing the image-to-image technique.}
  The input image undergoes a right shift transformation. The resulting image enters the image-to-image process to eliminate transformation artifacts and update the semantic content in alignment with the provided prompt.
  }
  \label{fig:animation_pipeline}
\end{figure}

Image-to-video generation involves a series of iterative steps, encompassing four stages as illustrated in Fig. \ref{fig:animation_pipeline}. Our animation pipeline is based on the Deforum technique 
\cite{deforum}. It consists of a series of transformations applied to the scene:

\begin{enumerate}
\item Conversion of the image into a three-dimensional representation using a depth map;
\item Application of spatial transformations to the resulting scene to induce an animated effect;
\item Projection of the 2.5D scene back onto a 2D image;
\item Elimination of transformation defects and updating of semantics through image-to-image conversion techniques.
\end{enumerate}

The scene generation process involves defining depth values along the $z$-axis within the interval $[(z_{\text{near}}, z_{\text{far}})]$ in the coordinate system. Depth estimation utilizes either MiDaS \cite{ranftl2019midas} or AdaBins \cite{bhat2020adabins}.
The camera is characterized by the coordinates $(x, y, z)$ in the three-dimensional space, and the direction of view, which is set by three angles $(\alpha, \beta, \gamma)$. Thus, we set the trajectory of the camera motion using the dependencies $x = x(t)$, $y = y(t)$, $z = z(t)$, $\alpha = \alpha(t)$, $\beta = 
\beta(t)$, and $\gamma = \gamma(t)$. The camera's first-person motion trajectory includes perspective projection operations with the camera initially fixed at the origin and the scene at a distance of $z_{\text{near}}$. Then, we apply transformations by rotating points around axes passing through the scene's center and translating to this center. Acknowledging the limitations of a single-image-derived depth map, addressing distortions resulting from minor camera orientation deviations is crucial. Two essential considerations follow: adjusting scene position through infinitesimal transformations and employing the image-to-image approach after each transformation. The image-to-image technique facilitates the realization of seamless and semantically accurate transitions between frames, enhancing the aesthetic appeal of this approach. The examples of image-to-video generations are presented in the \nameref{sec:Image-to-Video-appendix} appendix section.

\subsection{Text-to-Video}\label{sec:T2V}

Based on the Kandinsky 3.0 model, we also created the text-to-video generation pipeline Kandinsky Video \cite{arkhipkin2023fusionframes}, consisting, in addition to a text encoder and an image decoder, of two models -- for keyframes generation and interpolation between them. Both models use the pretrained weights of Kandinsky 3.0 as the backbone. We have also implemented the publicly available text-to-video generation interface\footnote{\url{https://fusionbrain.ai/en/t2v/}} in the Fusionbrain.ai website, which we mentioned above (Section \ref{sec:demo_system}). Please refer to the main paper for additional details regarding the text-to-video model.

\section{Human evaluation}\label{sec:HumanEval}

Some examples of text-to-image generations can be found in the \nameref{sec:text-to-image-results-appendix} appendix section. To compare Kandinsky 3.0 and Kandinsky 3.1 with other well-known models, we have compiled a balanced set of 2.1K prompts in 21 categories. Using this set, we first performed several side-by-side (SBS) comparisons between different versions of the Kandinsky 3.0 and then selected the best version. We conducted three rounds of SBS comparisons involving 28 people to do this. Next, we conducted side-by-side comparisons of Kandinsky 3.0 with the Kandinsky 2.2 \cite{razzhigaev2023kandinsky}, SDXL \cite{podell2023sdxl} and DALL-E 3 \cite{dalle3} models. Each study involved 12 people who voted an average of 30,000 times in each SBS. For this purpose, we developed chatbot that showed one of 2.1K pairs of images.

Each person chose the best image according to two criteria: 
\begin{enumerate}
  \item Alignment between image content and prompt (text comprehension);
  \item Visual quality of the image.
\end{enumerate}

We compared the visual quality and text comprehension in total for all categories, and each category separately. The visual examples of \nameref{sec:comparison-examples-appendix} and \nameref{sec:text-to-image-sbs-appendix} are presented in the appendix.

\section{Limitations}

Even though the current system can generate high-quality realistic images and successfully cope with diverse and complex language concepts, there are ways for further improvement. Among them is the improvement of semantic coherence between the input text and the generated image due to more efficient use of the text encoder potential. Challenges still remain high-fidelity text generation and photorealistic faces, and physics-controlled scene (lighting, positioning, focus and exposure, etc.).

\section{Border Impacts and Ethical Considerations}

Generative models are an effective tool for creativity and media content creation. They are also of great importance for the development of artificial intelligence science. We made the code and the trained weights of the model available to promote openness in the scientific community and the development of technologies that improve people's lives. We have provided free access to the user-friendly interface for everyone on the Internet.

At the same time, we know that generative models can be leveraged for blackmail, fraud, disinformation, creating fakes, inciting hatred and enmity between people, for unscrupulous political, financial, and other purposes. We warn against using our model in this way and strongly disagree with such malicious applications. We consider it necessary to note that the result of using the generations of our model for unfair purposes is entirely the user's responsibility.

 Despite this, we made many efforts to make sure that the generated images didn't contain malicious, offensive, or insulting content. To this end, we cleaned the training dataset from samples marked as harmful/offensive/abusive and removed offensive textual prompts. While obvious queries, according to our tests, rarely generate abusive content, there is technically no guarantee that some carefully designed prompts may not yield undesirable content. Therefore, depending on the application, we recommend using additional classifiers to filter out unwanted content and use image/representation transformation methods adapted to a given application.

\section{Conclusion}

In this report we highlighted the most significant advantages of our new text-to-image generative model -- Kandinsky 3.0.
Improving the text encoder size and extending the main diffusion U-Net we achieved higher human evaluation scores in comparison with Kandinsky 2.2. It should be mentioned that both measured quality indicators -- text understanding and visual quality improved. 
Comparing with SDXL we achieved significantly higher scores for both indicators. 

We described the acceleration of our model using distillation (Kandinsky 3.1 model). This allowed us to increase the inference speed by 20 times and reduce the number of steps in the reverse diffusion process to 4 without decrease in quality. We also show that the Kandinsky 3.0 model can be successfully used in various applications -- inpainting/outpainting, image editing, image-to-video, and text-to-video.

\bibliographystyle{unsrt}
\bibliography{references}

\newpage
\phantomsection

\section*{A. Acknowledgements}

The authors would also like to extend their deepest gratitude to the following list of teams and persons, who made a significant contribution to Kandinsky 3.0 research and development:
\begin{itemize}
    \item Sber AI Research team: Nikolay Gerasimenko, Elizaveta Dakhova, Sofia Kirillova, Mikhail Shoytov, Zein Shaheen, Anastasia Yaschenko;
    \item Anton Razzhigaev from FusionBrain research team at AIRI;
    \item Konstantin Kulikov and his production team at Sber AI;
    \item Sergey Markov and his research teams at SberDevices;
    \item Polina Voloshina labelling team;
    \item ABC Elementary labelling team;
    \item TagMe labelling team;
    \item Tatyana Nikulina, Angelina Kuts, Anton Bukashkin and prompt engineering team;
    \item Arseniy Shakhmatov, Anastasia Lysenko, Sergey Nesteruk, Ilya Ryabov and Mikhail Martynov (ex-Sber AI).
\end{itemize}

Thanks to all of you for your valuable help, advice, and constructive criticism.
\newpage

\section*{B. Generation Examples}

\subsection*{Text-to-Image}\label{sec:text-to-image-results-appendix}

\begin{figure}[H]
    \centering
    \includegraphics[bb=0 0 500 830, scale=0.75]{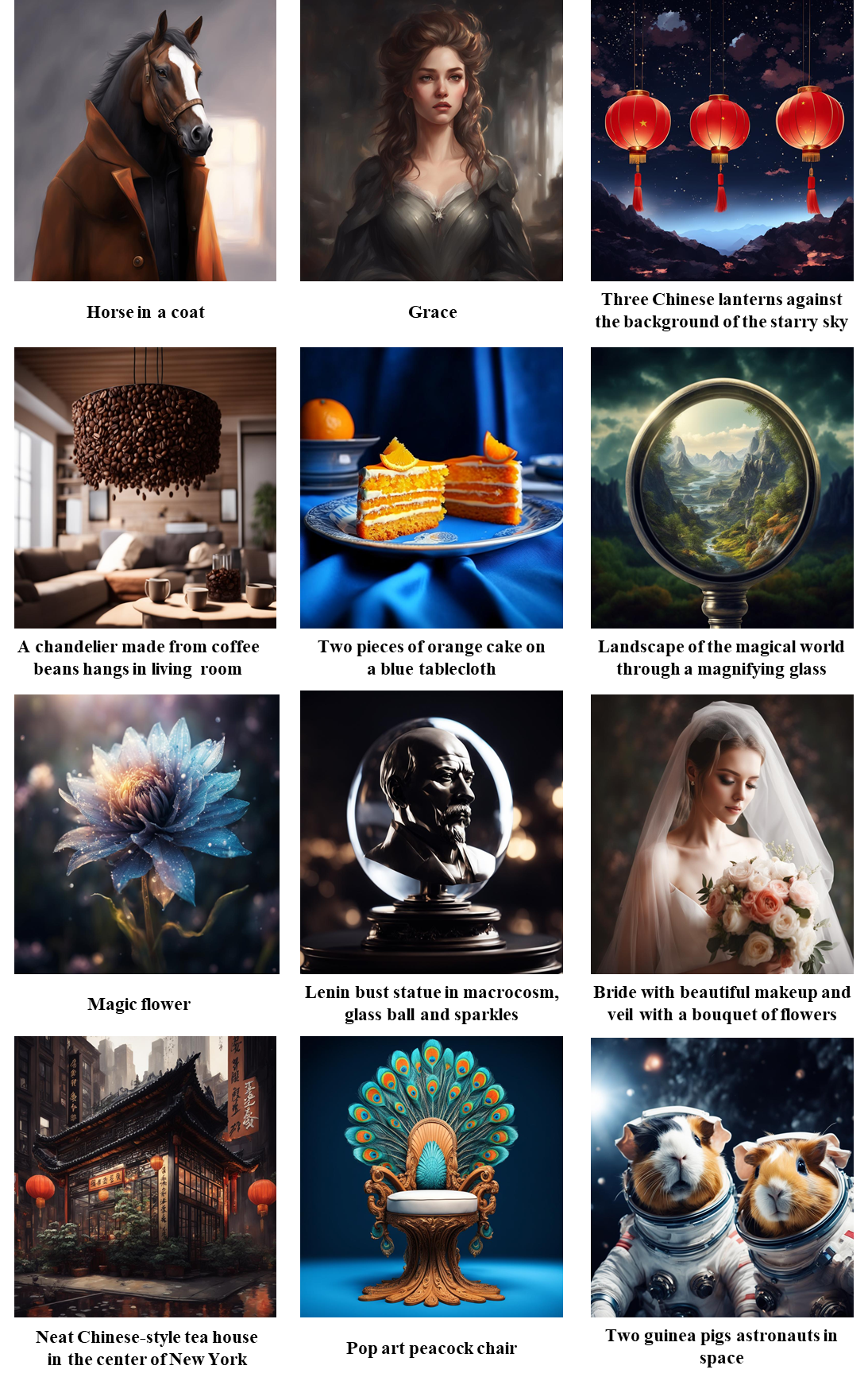}
    \label{fig:generation_images}
\end{figure}

\newpage

\subsection*{Distillation}\label{sec:Distillation-appendix}

\begin{figure}[H]
    \centering
    \includegraphics[bb=0 0 700 250, scale=0.7]{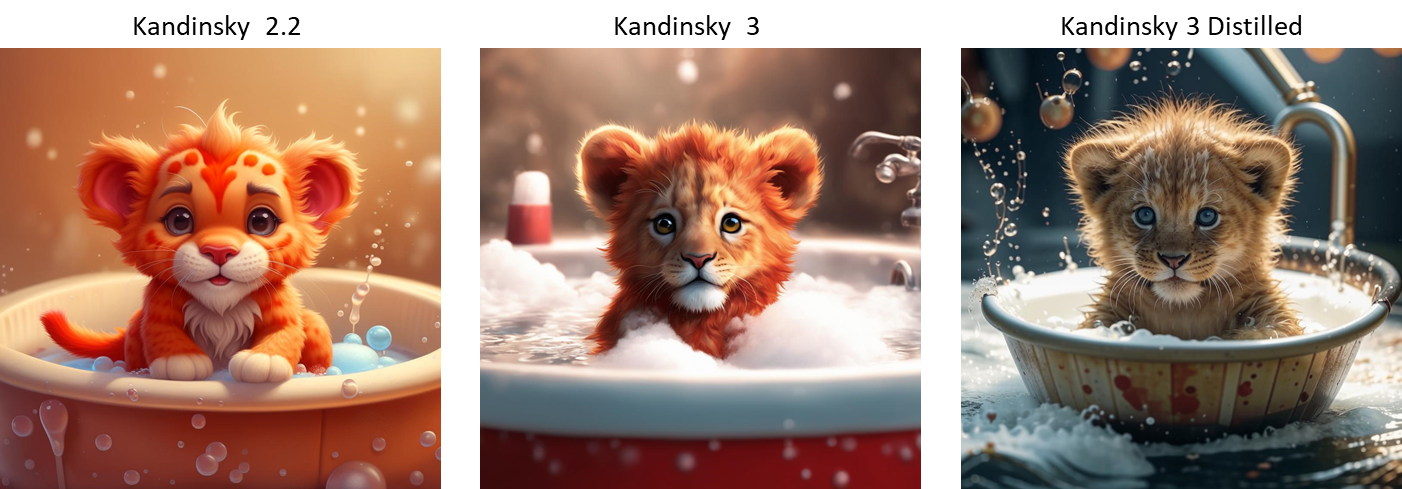}
    \label{fig:distillation1}
    \caption{\centering{{\textbf{Prompt:} \texttt{Cute red lion cub in a bath with foam}.}}}
\end{figure}

\begin{figure}[H]
    \centering
    \includegraphics[bb=0 0 700 230, scale=0.7]{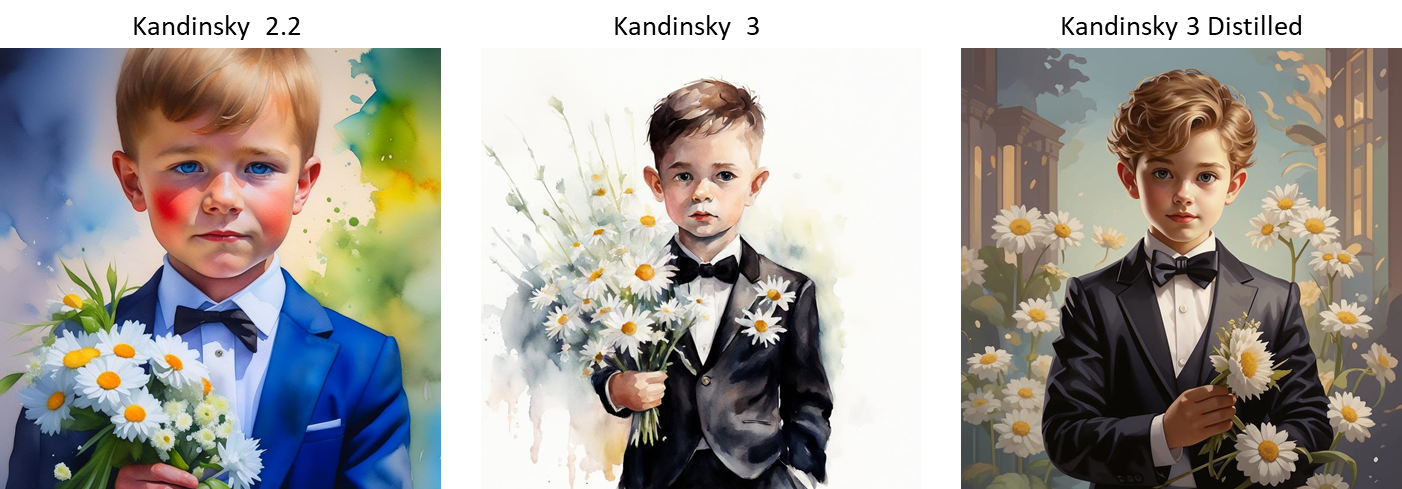}
    \label{fig:distillation2}
    \caption{\centering{{\textbf{Prompt:} \texttt{Close-up of a little boy in a tuxedo with a bouquet of white daisies, watercolor painting}.}}}
\end{figure}

\begin{figure}[H]
    \centering
    \includegraphics[bb=0 0 750 200, scale=0.7]{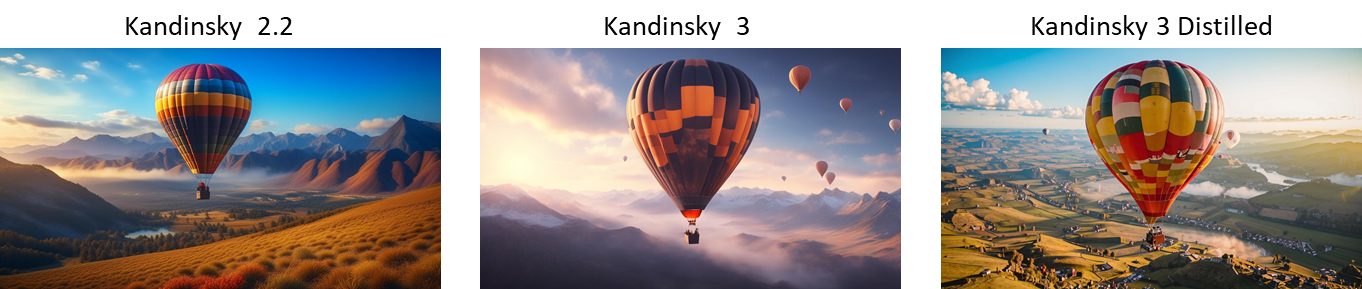}
    \label{fig:distillation3}
    \caption{\centering{{\textbf{Prompt:} \texttt{Huge hot air balloon soars in the fabulous steppes and mountain ranges, freshness, high image quality}.}}}
\end{figure}

\begin{figure}[H]
    \centering
    \includegraphics[bb=0 0 700 350, scale=1]{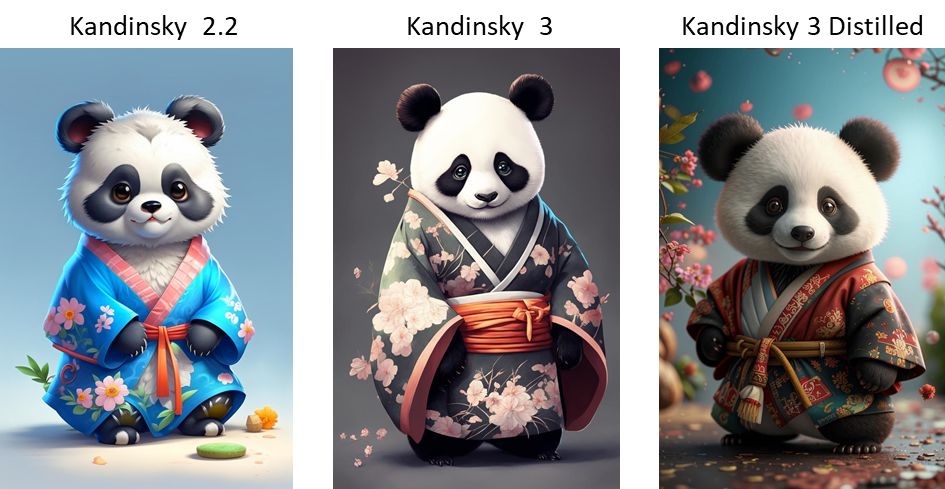}
    \label{fig:distillation4}
    \caption{\centering{{\textbf{Prompt:} \texttt{Cute little panda stands in a kimono, a very kind cute look, cute eyes, kindness}.}}}
\end{figure}

\newpage

\subsection*{Prompt beautification}\label{sec:PromptBeautification}

\begin{figure}[H]
    \centering
    \includegraphics[bb=0 0 410 200, scale=0.8]{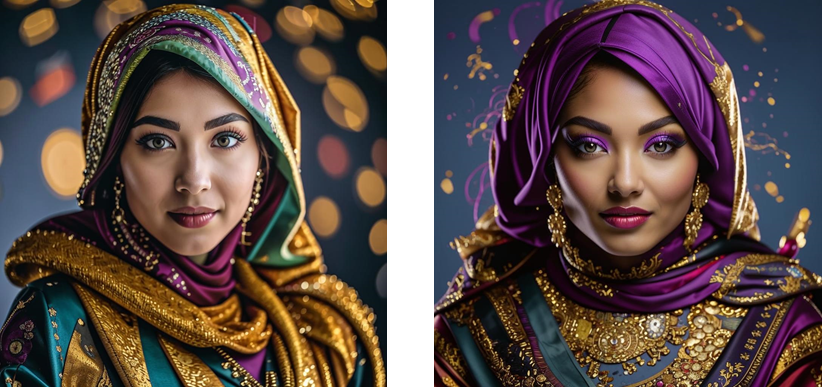}
    \label{fig:beautification1}
    \caption{\centering{{\textbf{Prompt:} \texttt{Close-up photo of a beautiful oriental woman, elegant hijab-adorned with hints of modern vintage style}. Without/With LLM.}}}
\end{figure}

\begin{figure}[H]
    \centering
    \includegraphics[bb=0 0 410 200, scale=0.8]{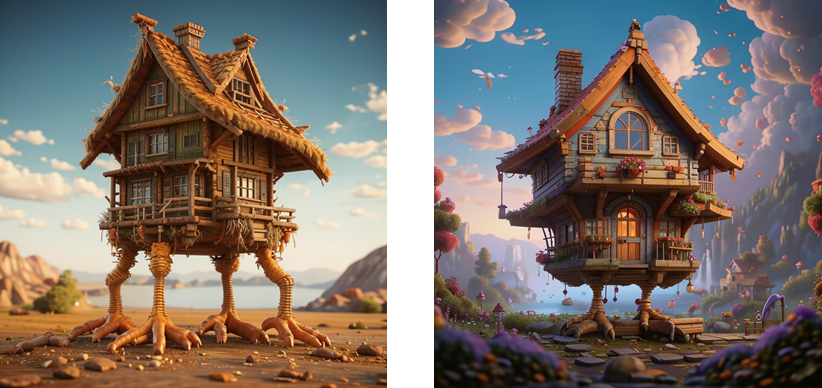}
    \label{fig:beautification2}
    \caption{\centering{{\textbf{Prompt:} \texttt{A hut on chicken legs}. Without/With LLM.}}}
\end{figure}

\begin{figure}[H]
    \centering
    \includegraphics[bb=0 0 700 200, scale=0.7]{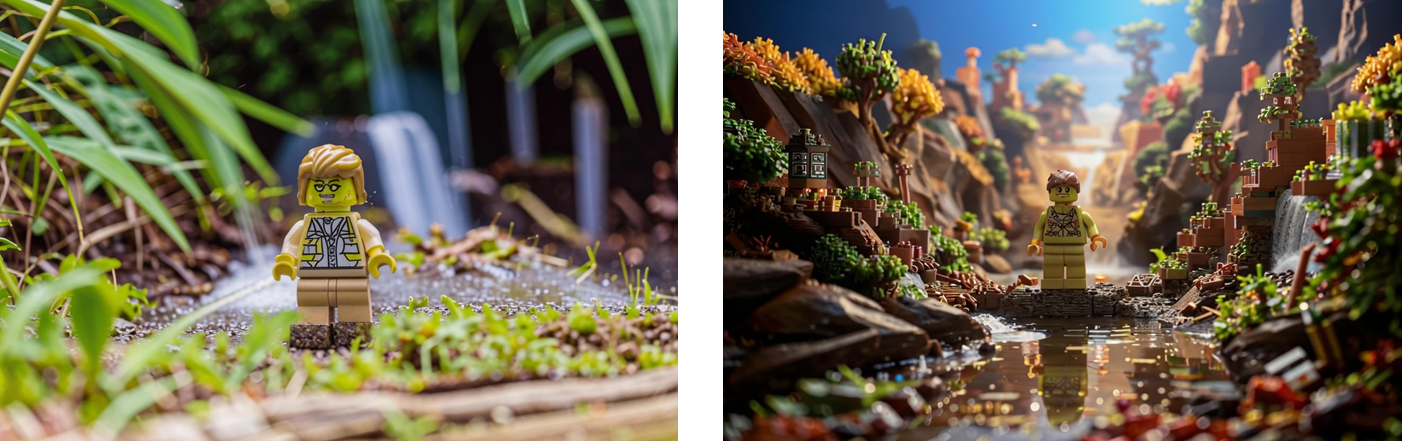}
    \label{fig:beautification3}
    \caption{\centering{{\textbf{Prompt:} \texttt{Lego figure at the waterfall}. Without/With LLM.}}}
\end{figure}

\newpage

\subsection*{Outpainting}\label{sec:outpainting_appendix}
        \begin{figure}[H]
            \centering
                \begin{tabular}{ll}
                   \includegraphics[bb=0 0 1024 1024, width=0.4\linewidth]{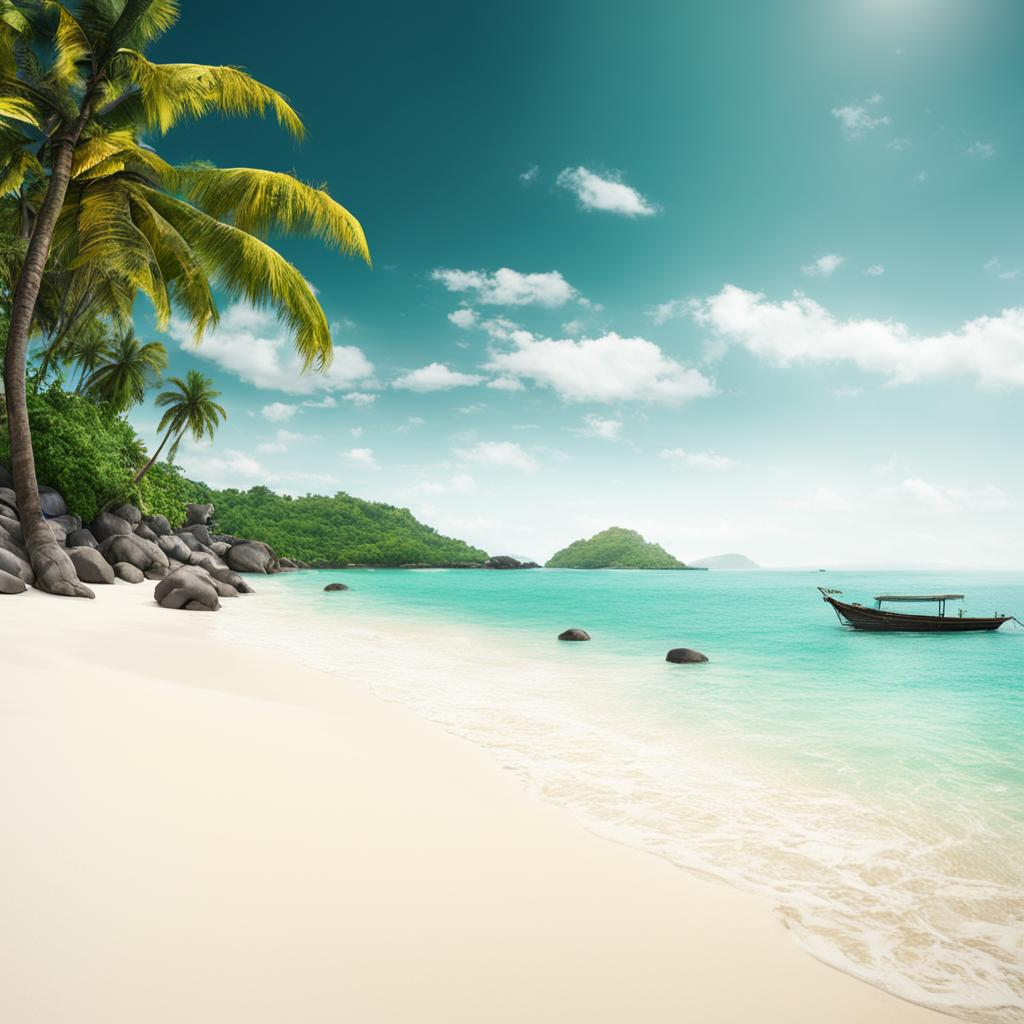}
                    & \includegraphics[bb=0 0 1536 1024, width=0.6\linewidth]{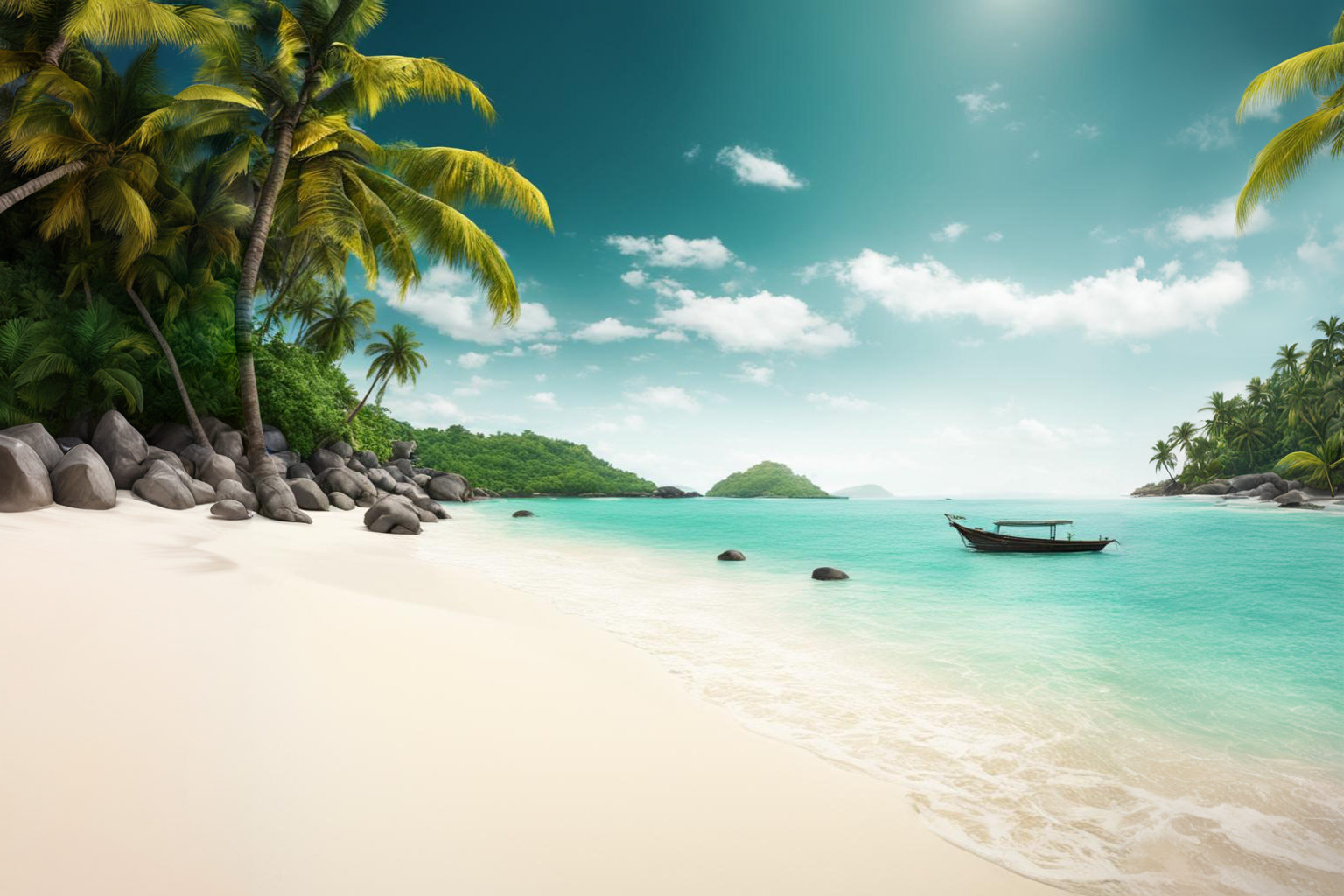}\\ 
                    \includegraphics[bb=0 0 1024 1024, width=0.4\linewidth]{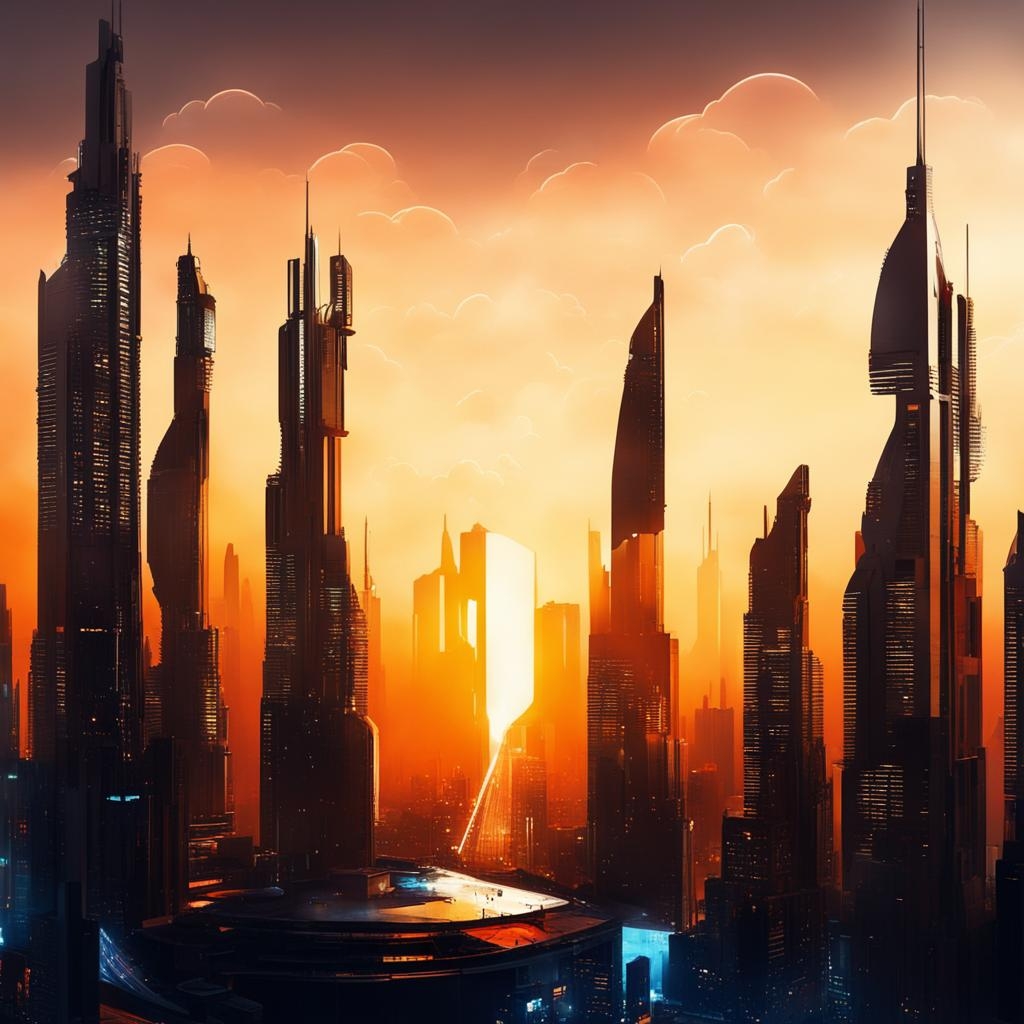} & \includegraphics[bb=0 0 1535 1024, width=0.6\linewidth]{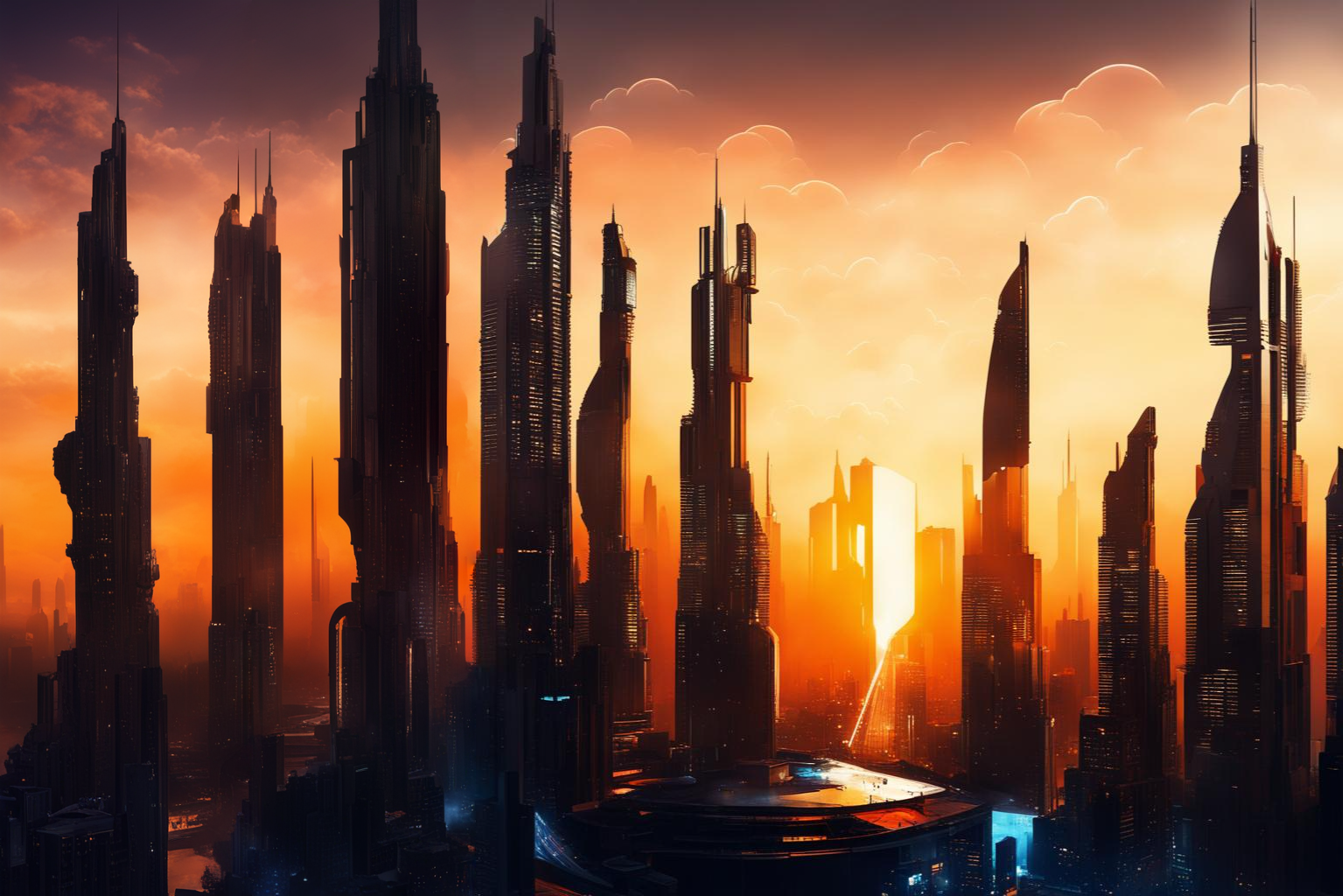}\\
                     \includegraphics[bb=0 0 1024 1024, width=0.4\linewidth]{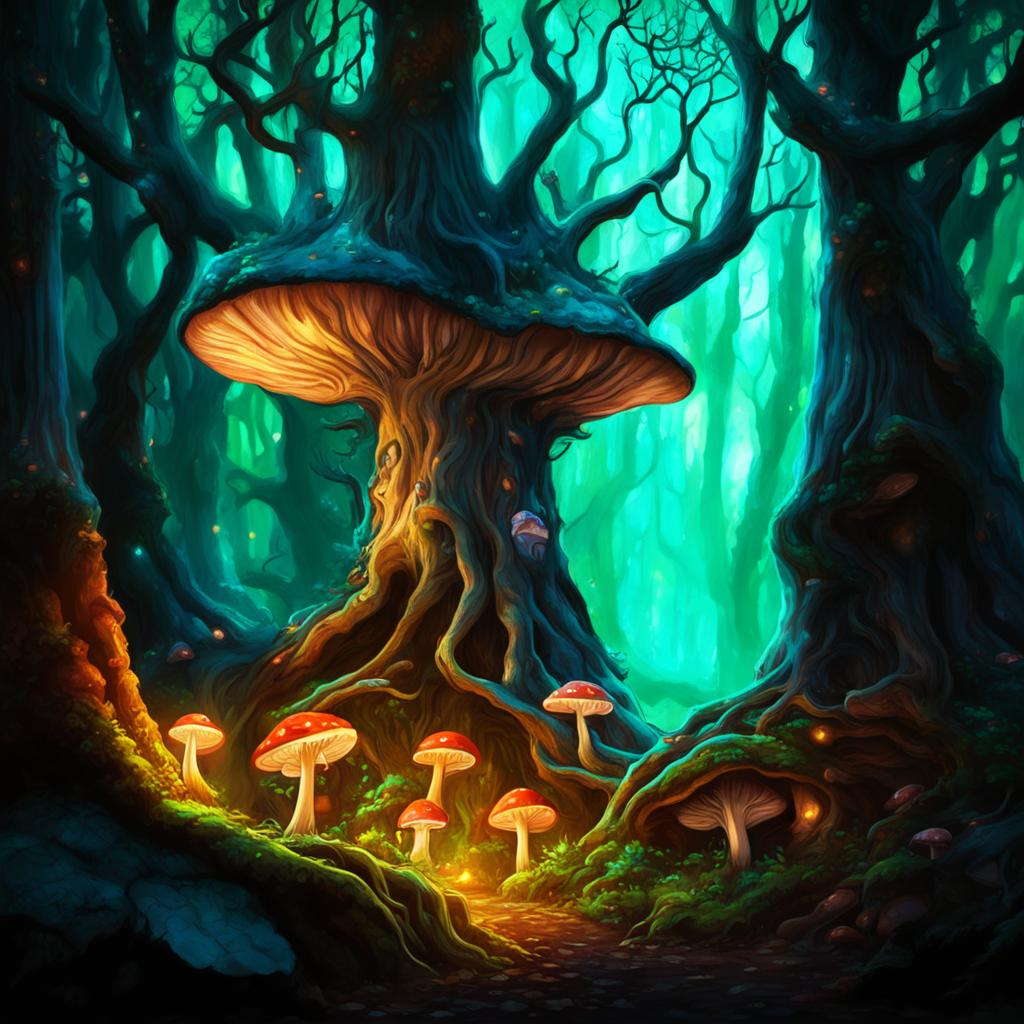}
                    & \includegraphics[bb=0 0 1536 1024, width=0.6\linewidth]{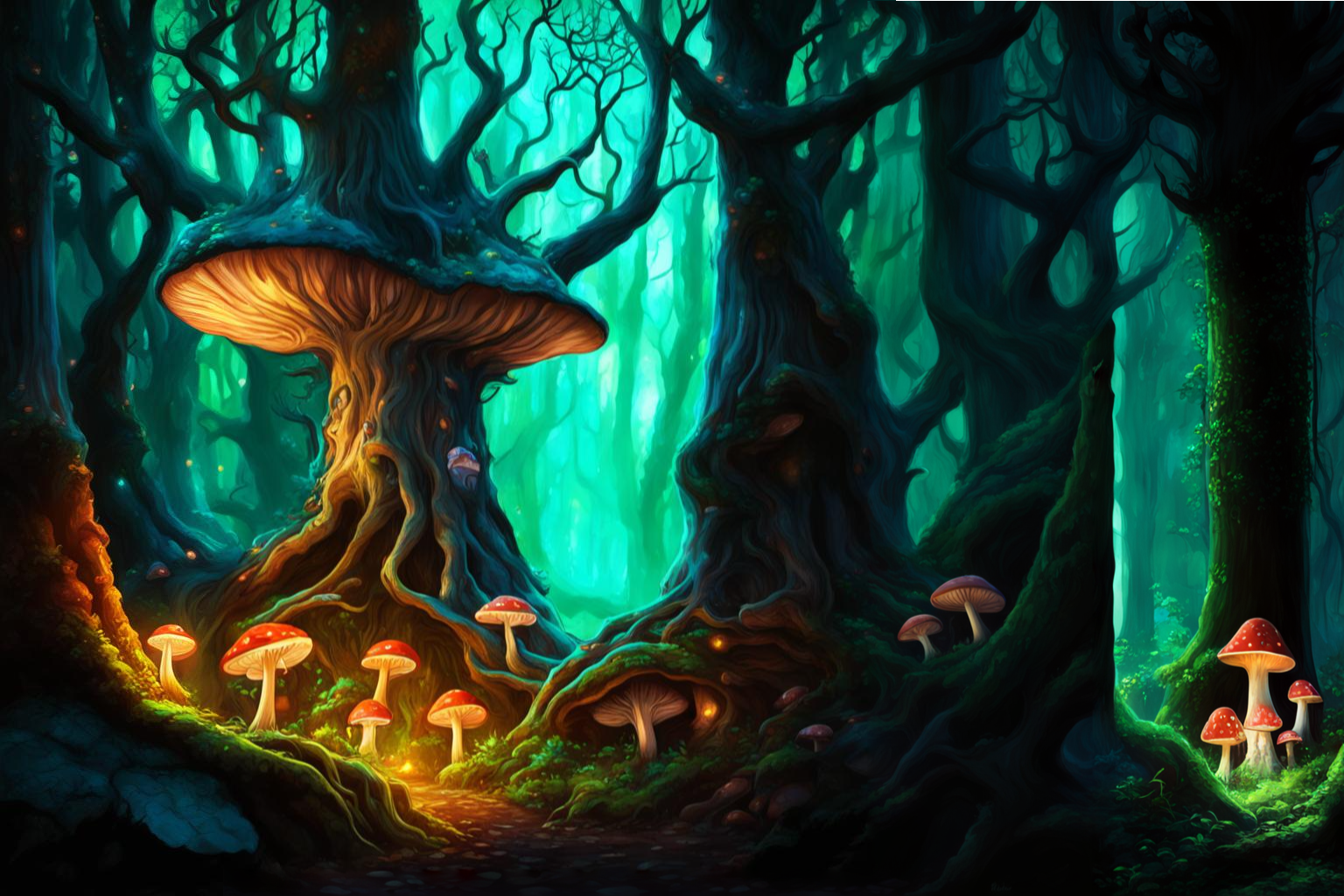}\\
                \end{tabular}%
             \caption{Examples of outpainting.\label{fig:outpainting}}
        \end{figure}

\newpage

\subsection*{Image Editing with IP-Adapter}\label{sec:IP-Adapter-appendix}

\begin{figure}[H]
    \centering
    \includegraphics[bb=0 0 800 200, scale=0.6]{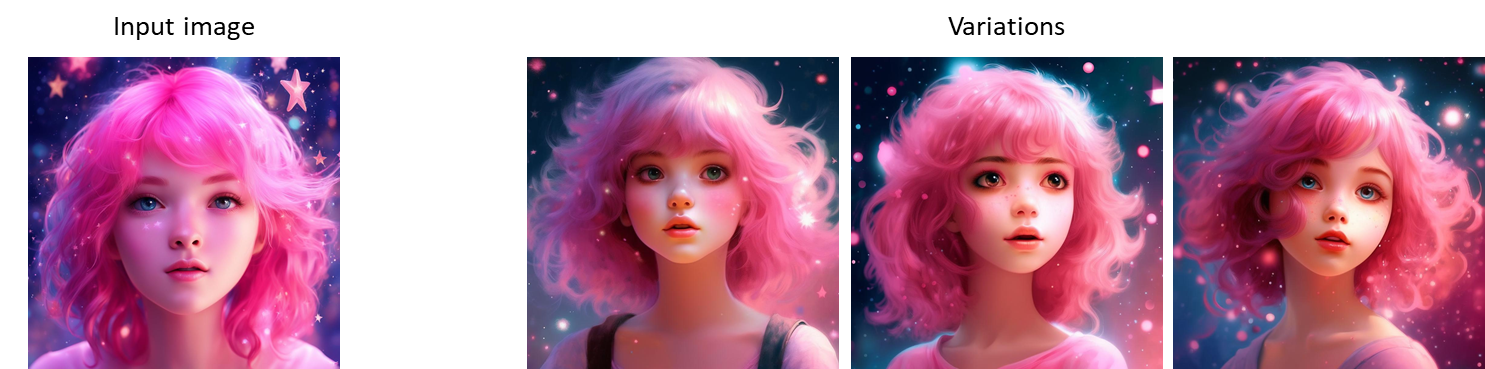}
    \caption{}
\end{figure}

\begin{figure}[H]
    \centering
    \includegraphics[bb=0 0 600 200, scale=0.55]{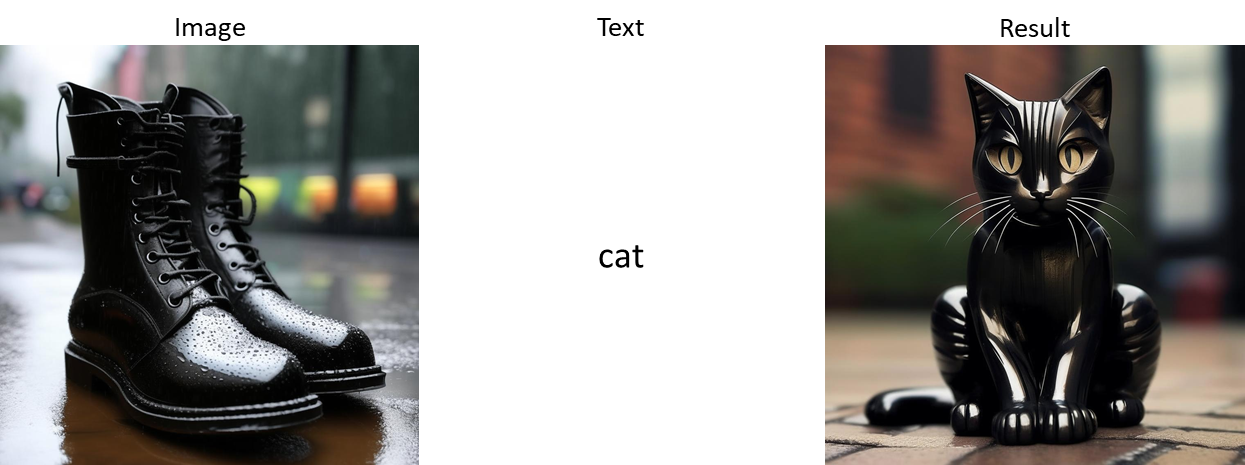}
    \caption{}
\end{figure}

\begin{figure}[H]
    \centering
    \includegraphics[bb=0 0 600 200, scale=0.55]{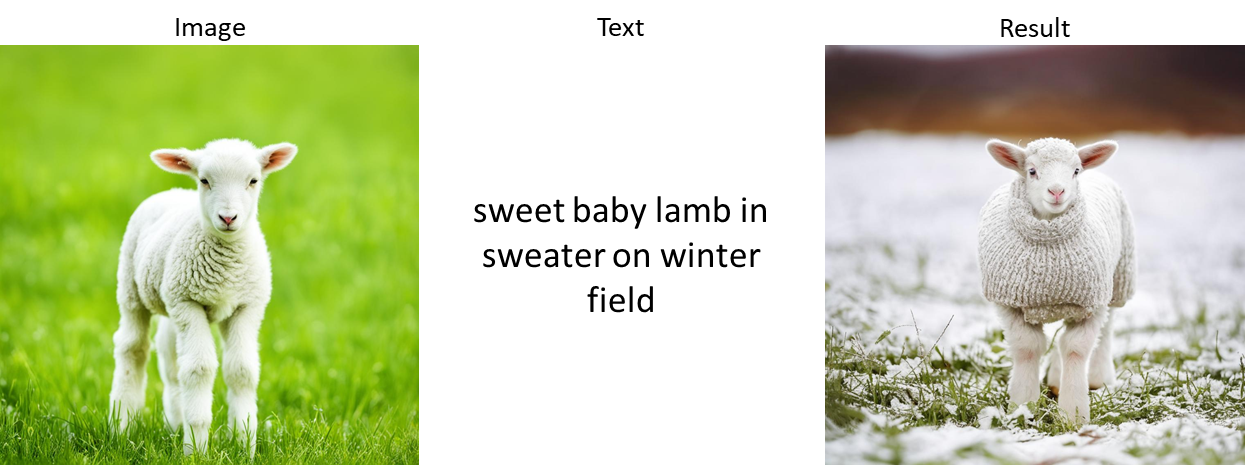}
    \caption{}
\end{figure}

\begin{figure}[H]
    \centering
    \includegraphics[bb=0 0 570 200, scale=0.6]{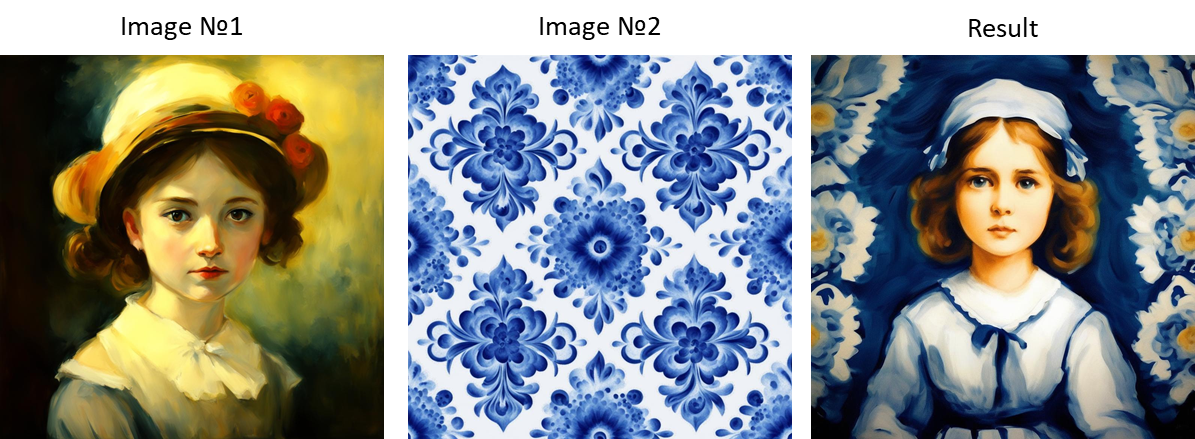}
    \caption{}
\end{figure}

\newpage

\subsection*{Image-to-Video}\label{sec:Image-to-Video-appendix}

\begin{figure}[H]
    \center{\includegraphics[bb=0 0 722 587, scale=0.7
    ]{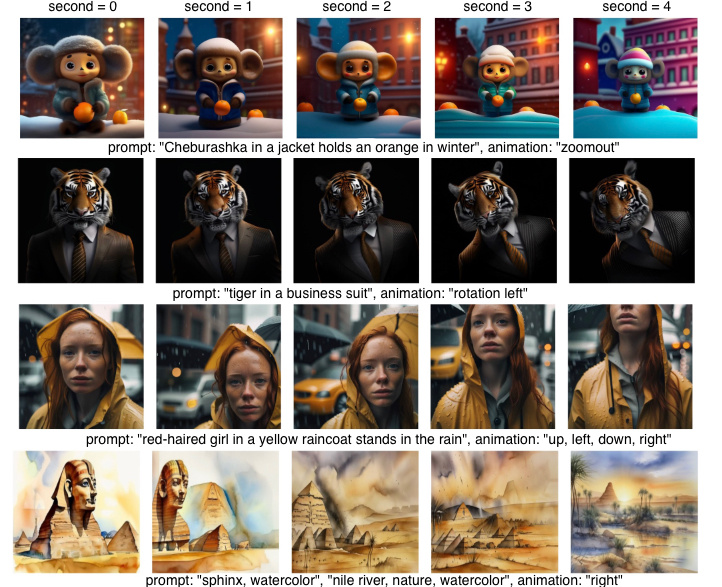}}
    \caption{\centering{{Animations generated by Image-to-Video pipeline.}}}
\end{figure}

\newpage

\subsection*{Comparison to prior works}\label{sec:comparison-examples-appendix}




\begin{figure}[H]
    \center{\includegraphics[bb=0 0 700 370, scale=0.7
    ]{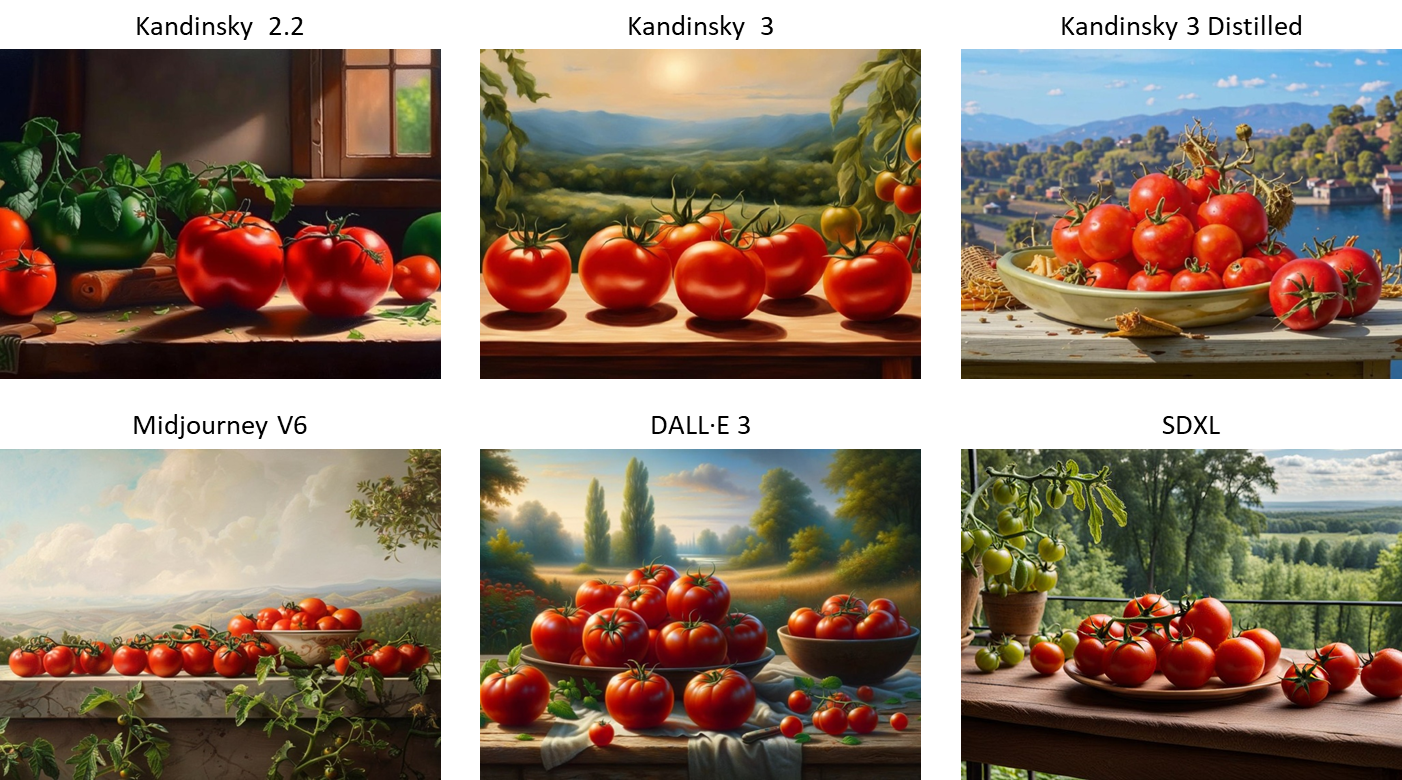}}
    \caption{\centering{{\textbf{Prompt:} \texttt{Tomatoes on a table, against the backdrop of nature, a still life painting depicted in a hyper realistic style}.}}}
    \label{fig:from_sbs1}
\end{figure}

\begin{figure}[H]
    \center{\includegraphics[bb=0 0 700 370, scale=0.7
    ]{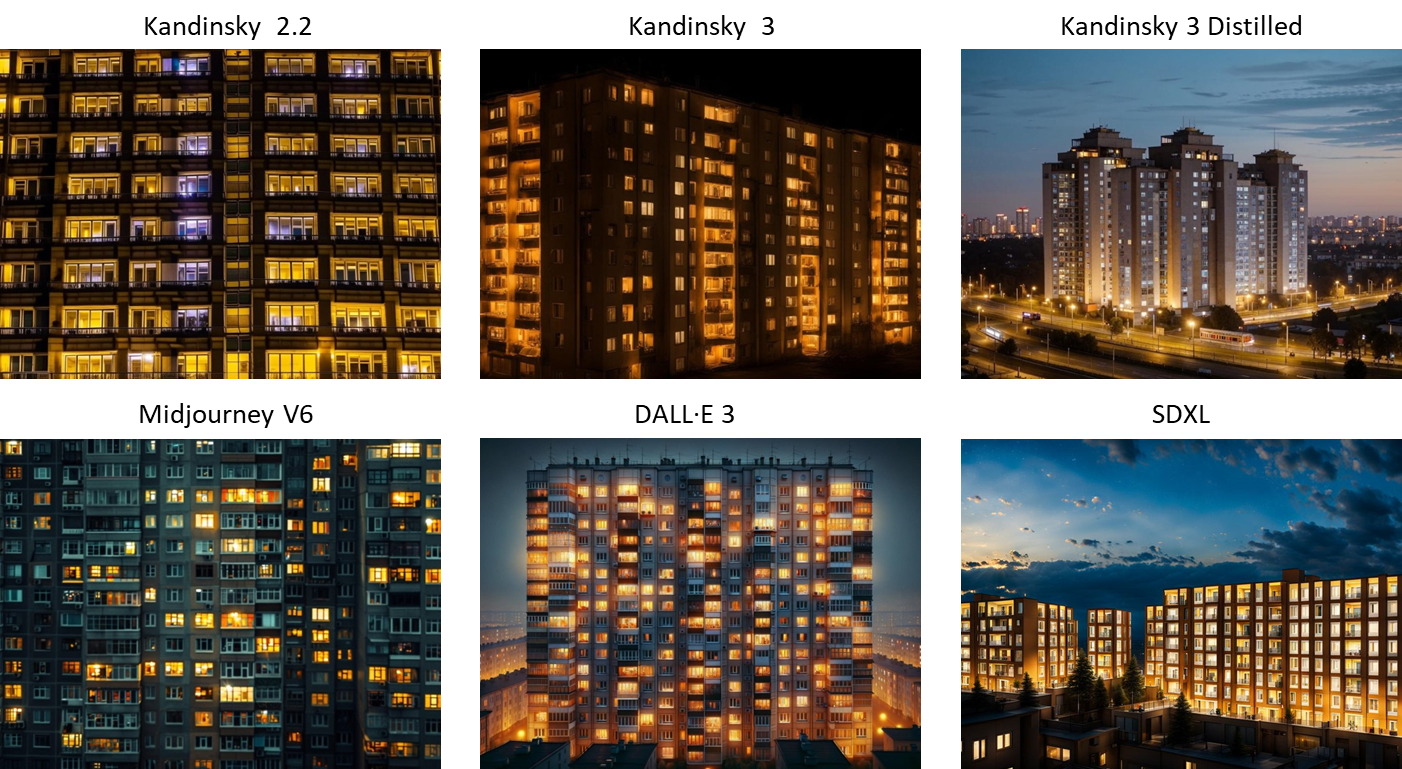}}
    \caption{\centering{{\textbf{Prompt:} \texttt{Photo of Russian apartment buildings with many windows at night, lights on inside each window, the building is lighted from behind by warm yellow and white lights, it's a wide shot showing an entire multistorey block of flats, there are more houses in front of us, high resolution, hyper realistic}.}}}
    \label{fig:from_sbs2}
\end{figure}

\begin{figure}[H]
    \center{\includegraphics[bb=0 0 555 370, scale=0.7
    ]{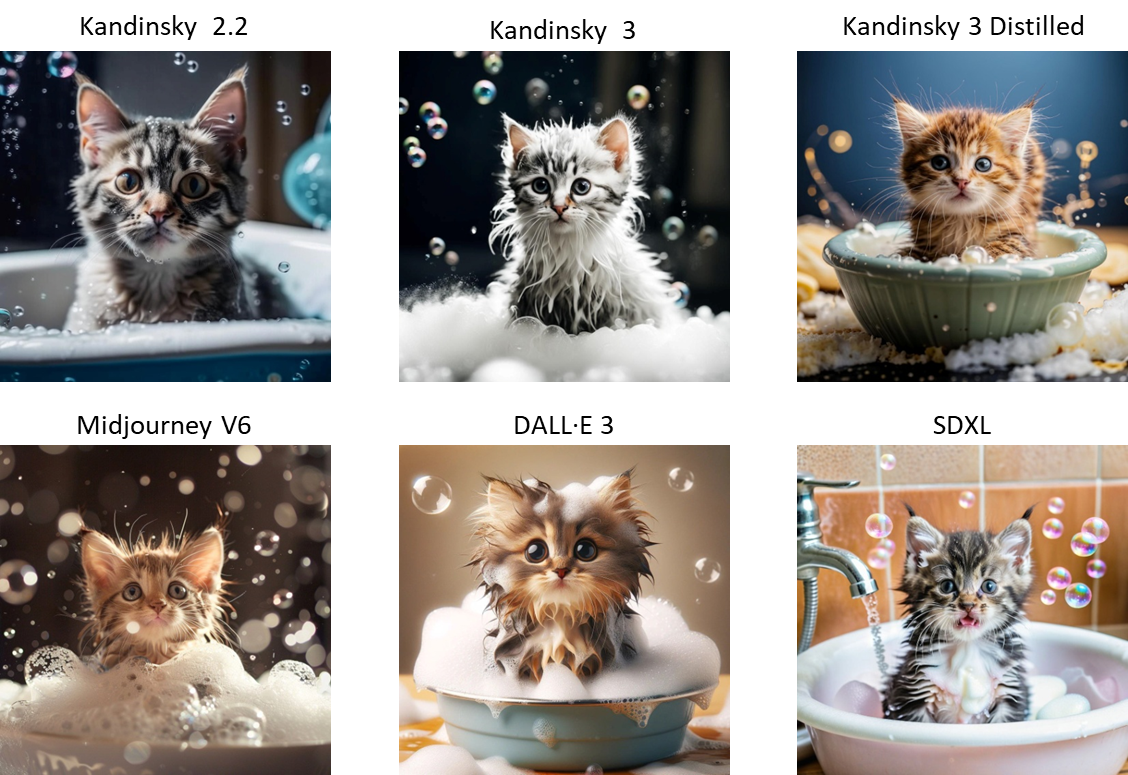}}
    \caption{\centering{{\textbf{Prompt:} \texttt{Funny cute wet kitten sitting in a basin with soap foam, soap bubbles around, photography}.}}}
    \label{fig:from_sbs3}
\end{figure}

\begin{figure}[H]
    \center{\includegraphics[bb=0 0 550 370, scale=0.7
    ]{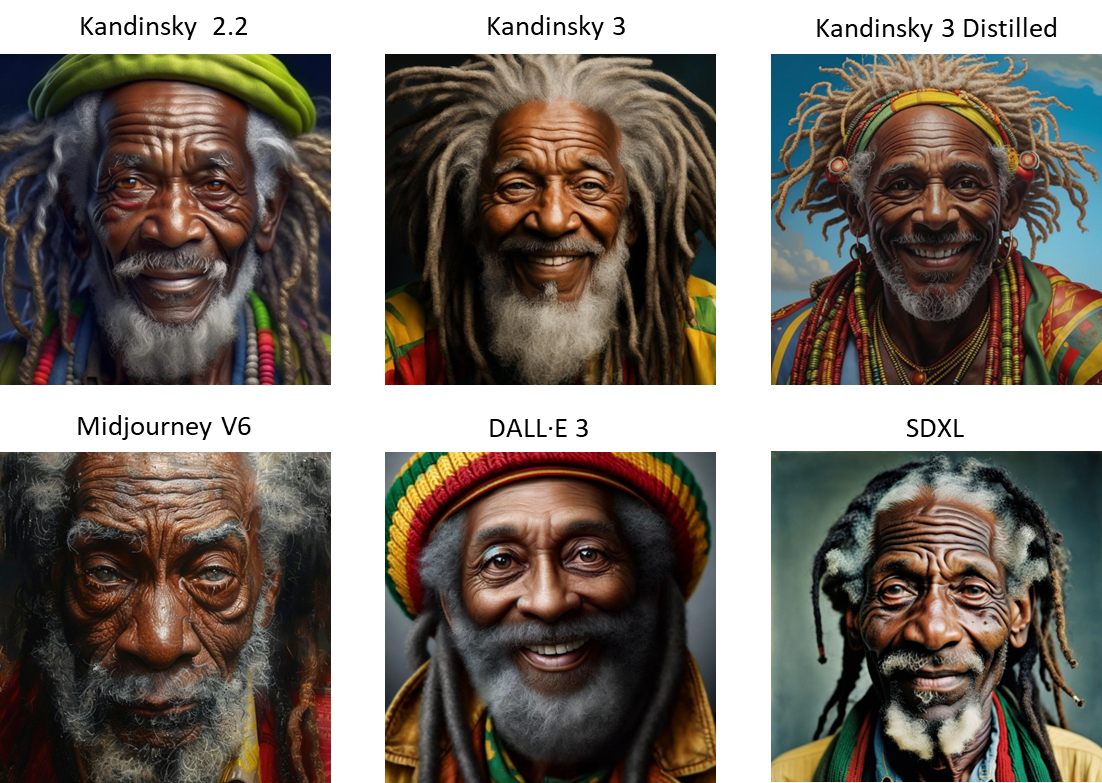}}
    \caption{\centering{{\textbf{Prompt:} \texttt{Maximally realistic portrait of a jolly old gray-haired Negro Rastaman with wrinkles around his eyes and a crooked nose in motley clothes}.}}}
    \label{fig:from_sbs4}
\end{figure}

\section*{C. Side-by-side human evaluation results}\label{sec:SBS}

\subsection*{Human evaluation results for text-to-image generation}\label{sec:text-to-image-sbs-appendix}

\begin{figure}[H]
    \center{\includegraphics[bb=0 0 1000 570, scale=0.44
    ]{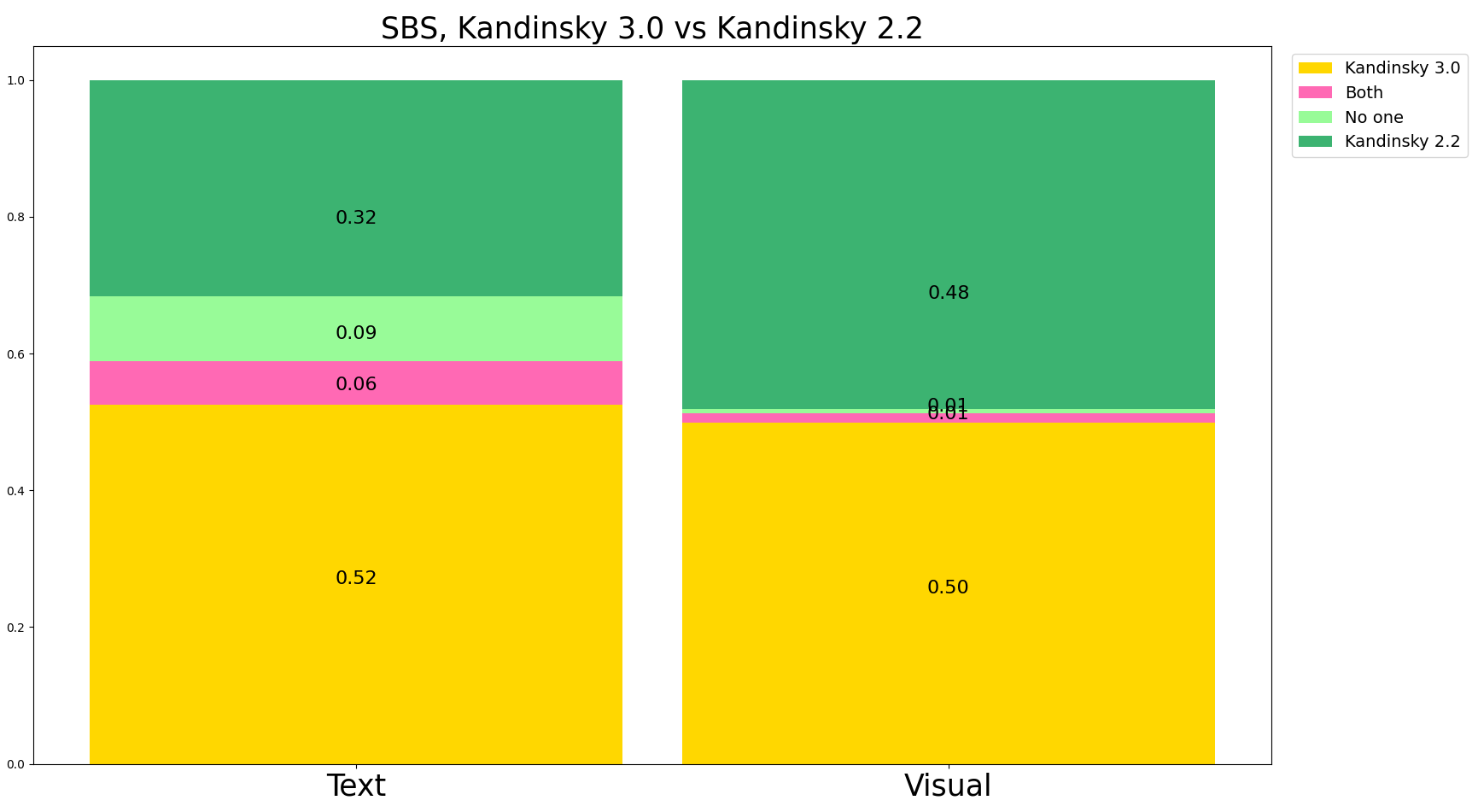}}
    \caption{\centering{Overall results of side-by-side human comparison between Kandinsky 3.0 and Kandinsky 2.2.}}
\end{figure}

\begin{figure}[H]
    \center{\includegraphics[bb=0 0 1000 570, scale=0.44
    ]{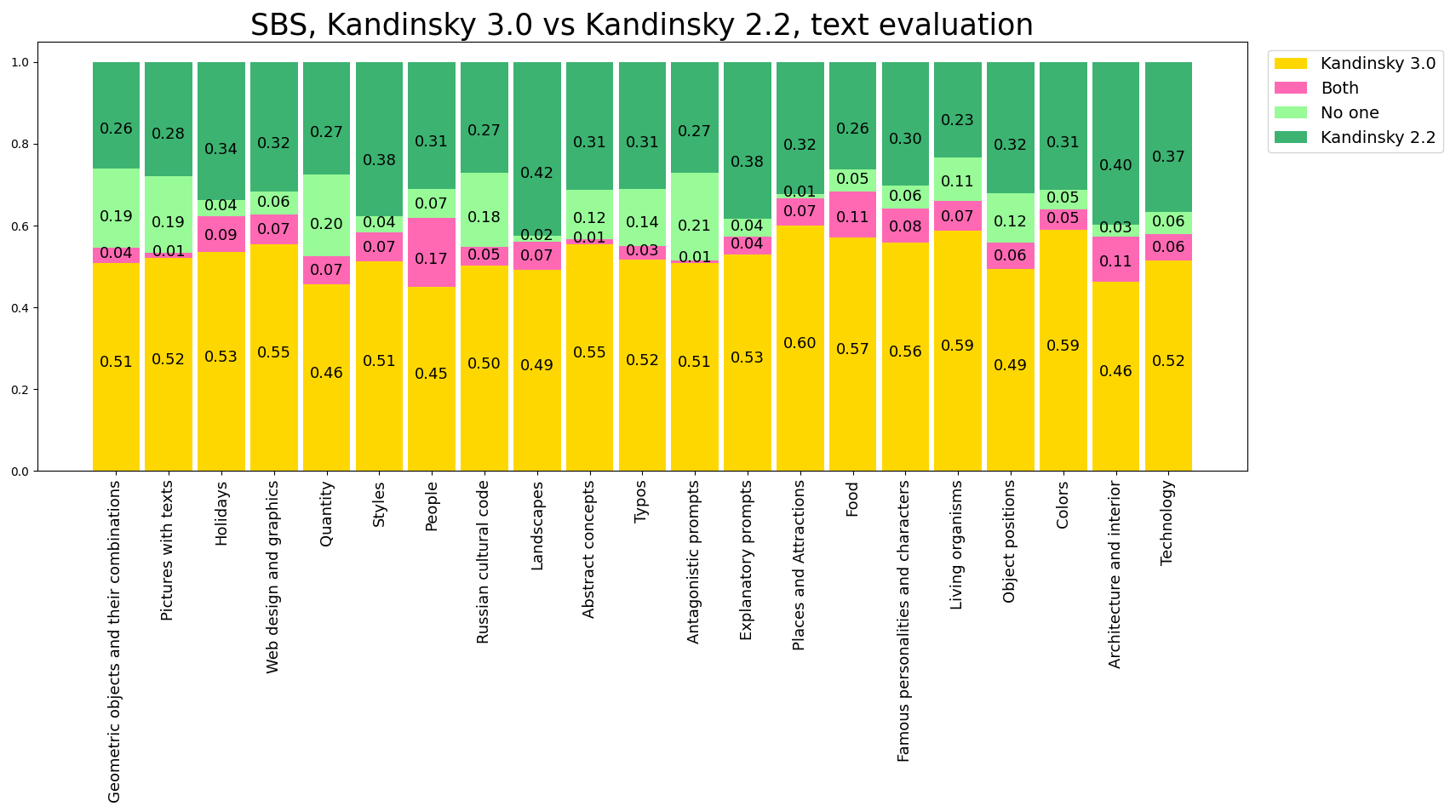}}
    \caption{\centering{Results of side-by-side human comparison between Kandinsky 3.0 and Kandinsky 2.2 for \textbf{text comprehension}.}}
    \label{fig:kandy3_kandy2_text}
\end{figure}

\begin{figure}[H]
    \center{\includegraphics[bb=0 0 1000 570, scale=0.44
    ]{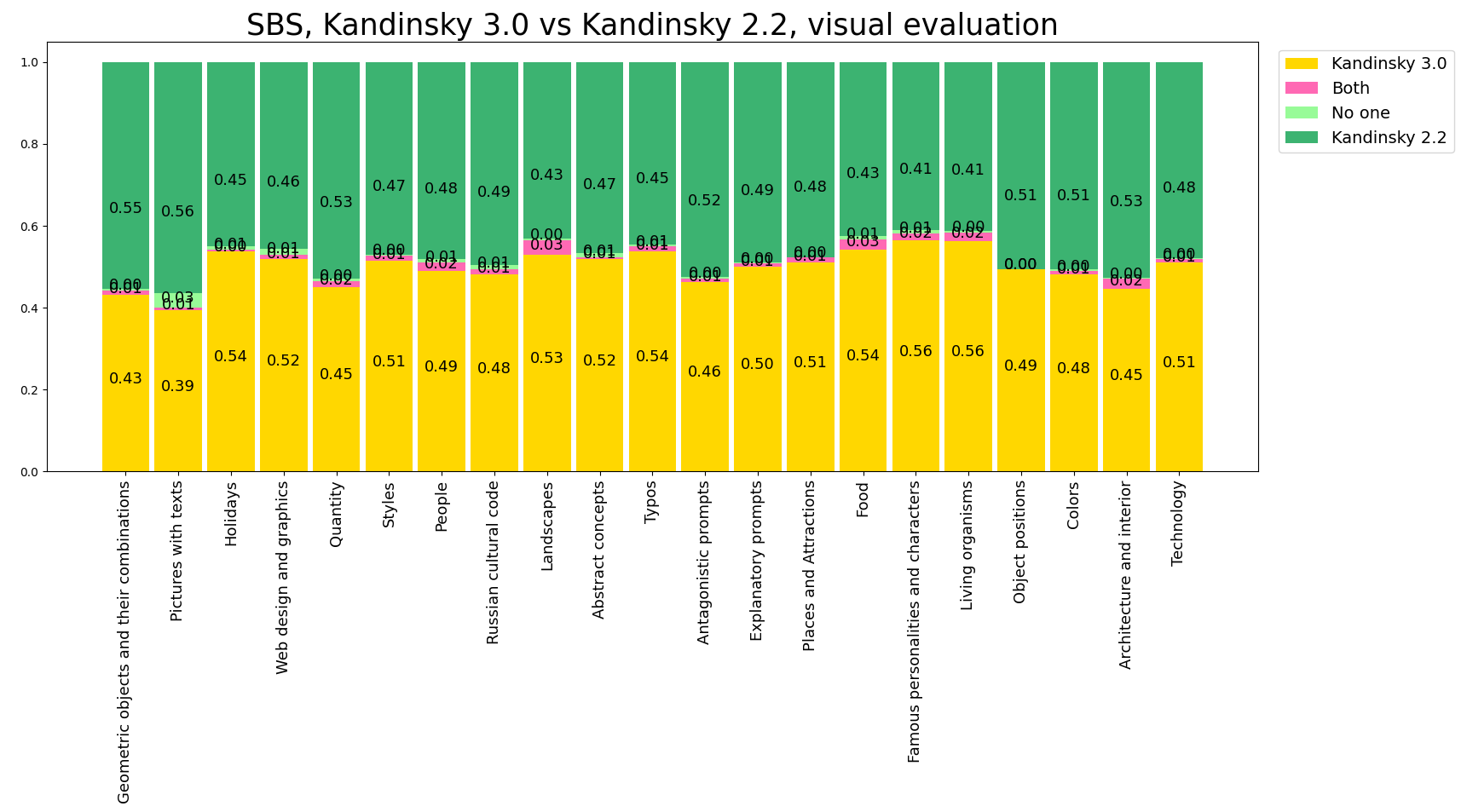}}
    \caption{\centering{Results of side-by-side human comparison between Kandinsky 3.0 and Kandinsky 2.2 for \textbf{visual quality}.}}
    \label{fig:kandy3_kandy2_visual}
\end{figure}

\begin{figure}[H]
    \center{\includegraphics[bb=0 0 1000 570, scale=0.44
    ]{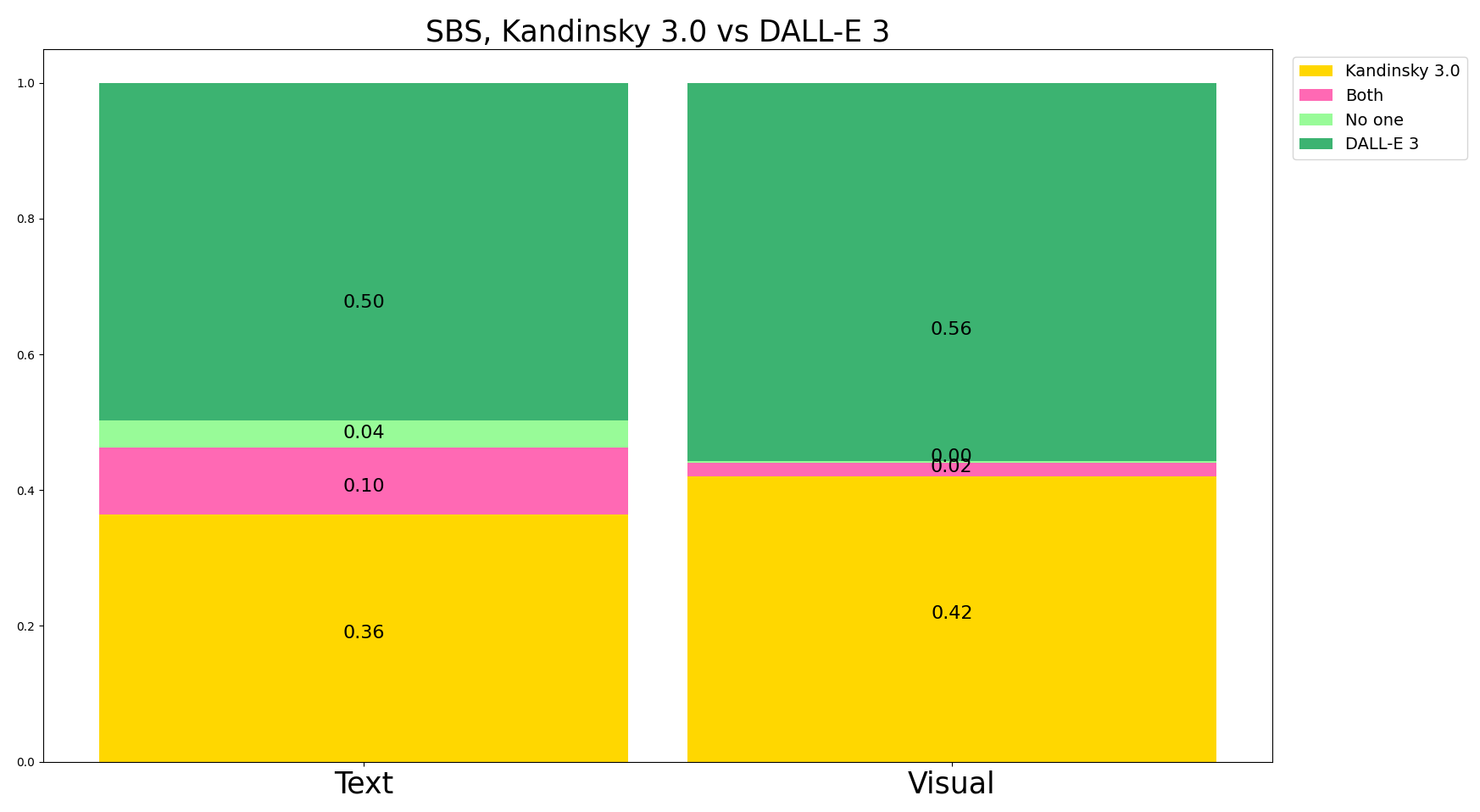}}
    \caption{\centering{Overall results of side-by-side human comparison between Kandinsky 3.0 and DALL-E 3.}}
    \label{fig:kandy3_dalle3_all}
\end{figure}

\begin{figure}[H]
    \center{\includegraphics[bb=0 0 1000 570, scale=0.45
    ]{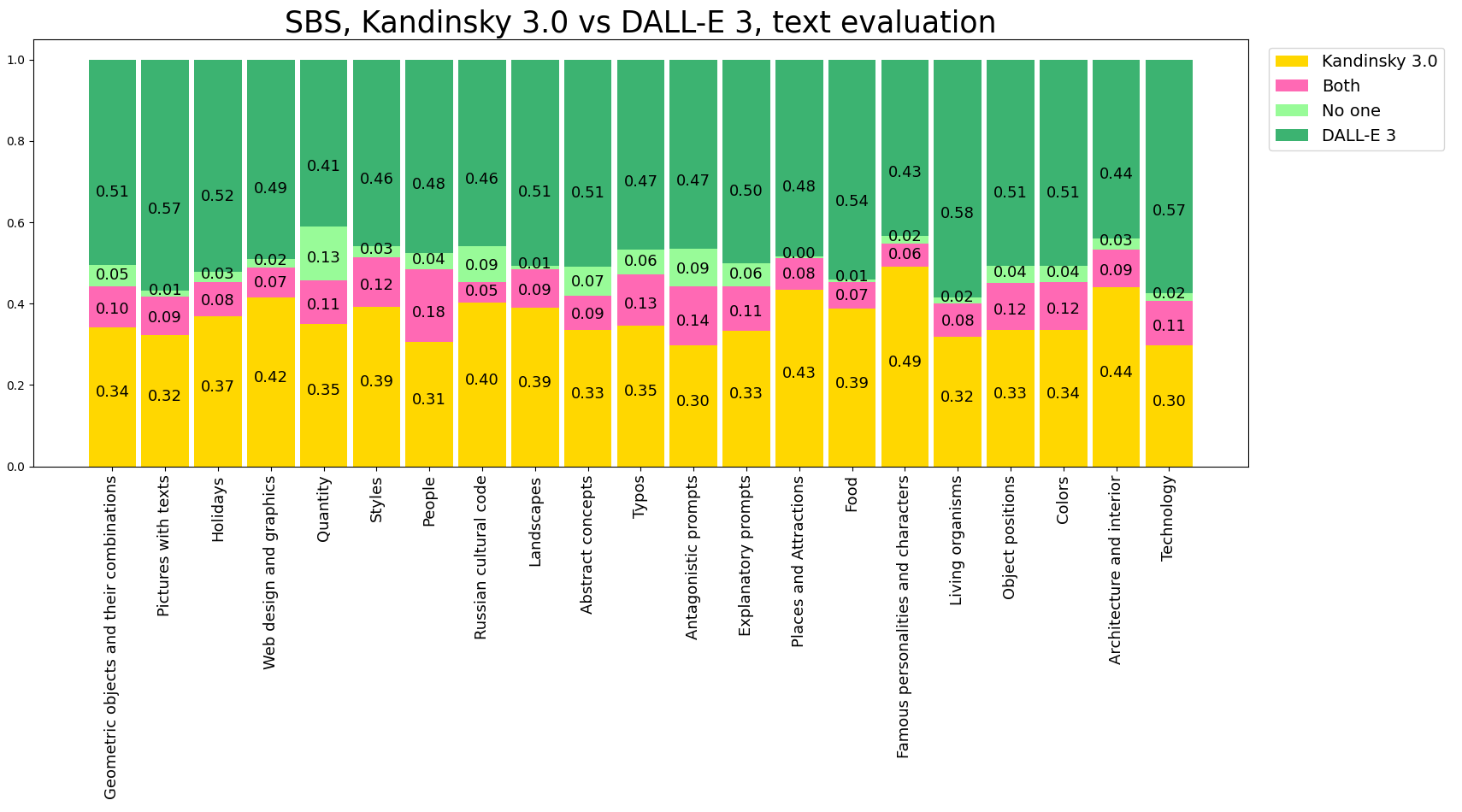}}
    \caption{\centering{Results of side-by-side human comparison between Kandinsky 3.0 and DALL-E 3 for \textbf{text comprehension}.}}
    \label{fig:kandy3_dalle3_text}
\end{figure}

\begin{figure}[H]
    \center{\includegraphics[bb=0 0 1000 570, scale=0.45
    ]{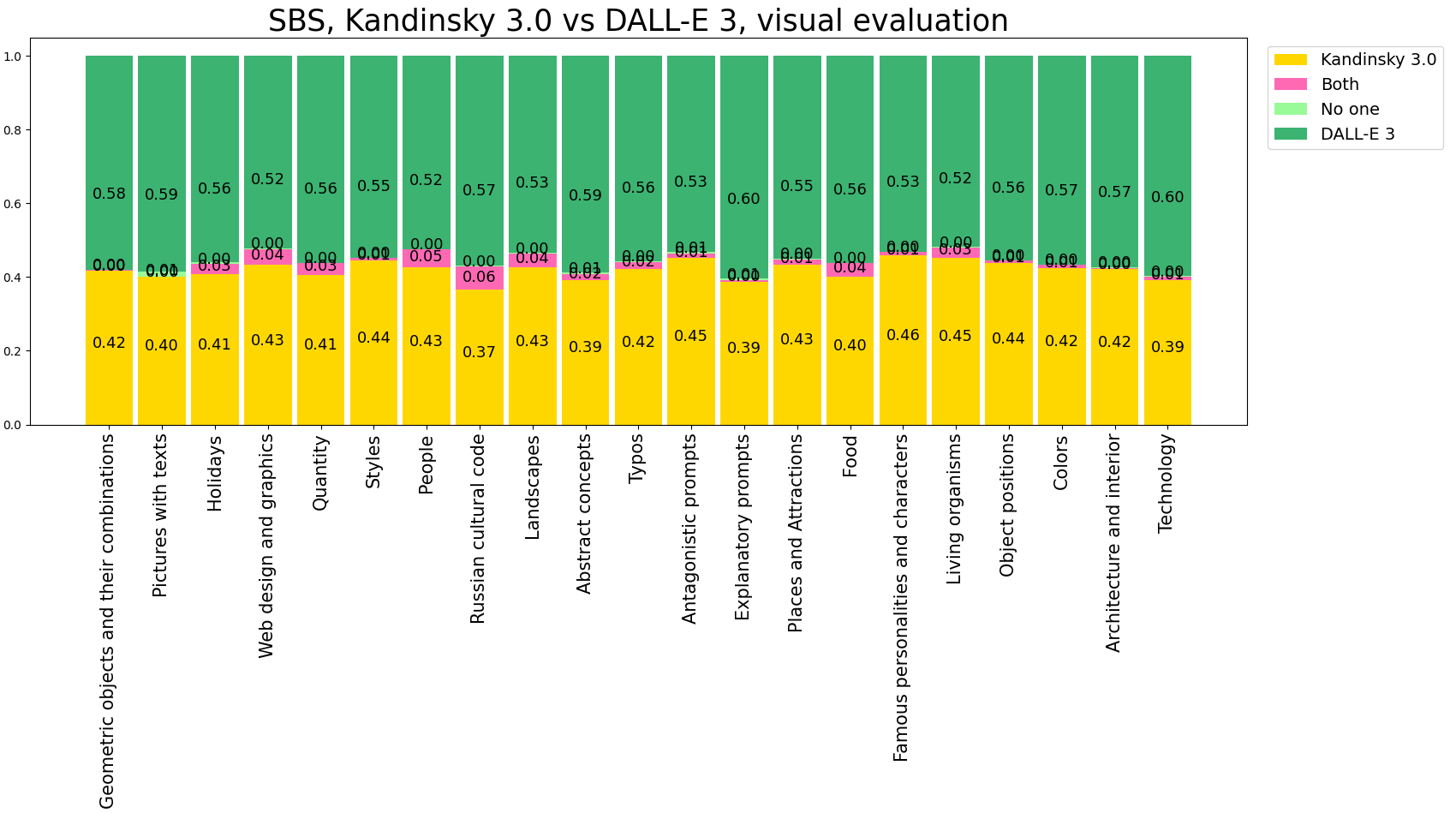}}
    \caption{\centering{Results of side-by-side human comparison between Kandinsky 3.0 and DALL-E 3 for \textbf{visual quality}.}}
    \label{fig:kandy3_dalle3_visual}
\end{figure}

\begin{figure}[H]
    \center{\includegraphics[bb=0 0 1000 570, scale=0.44
    ]{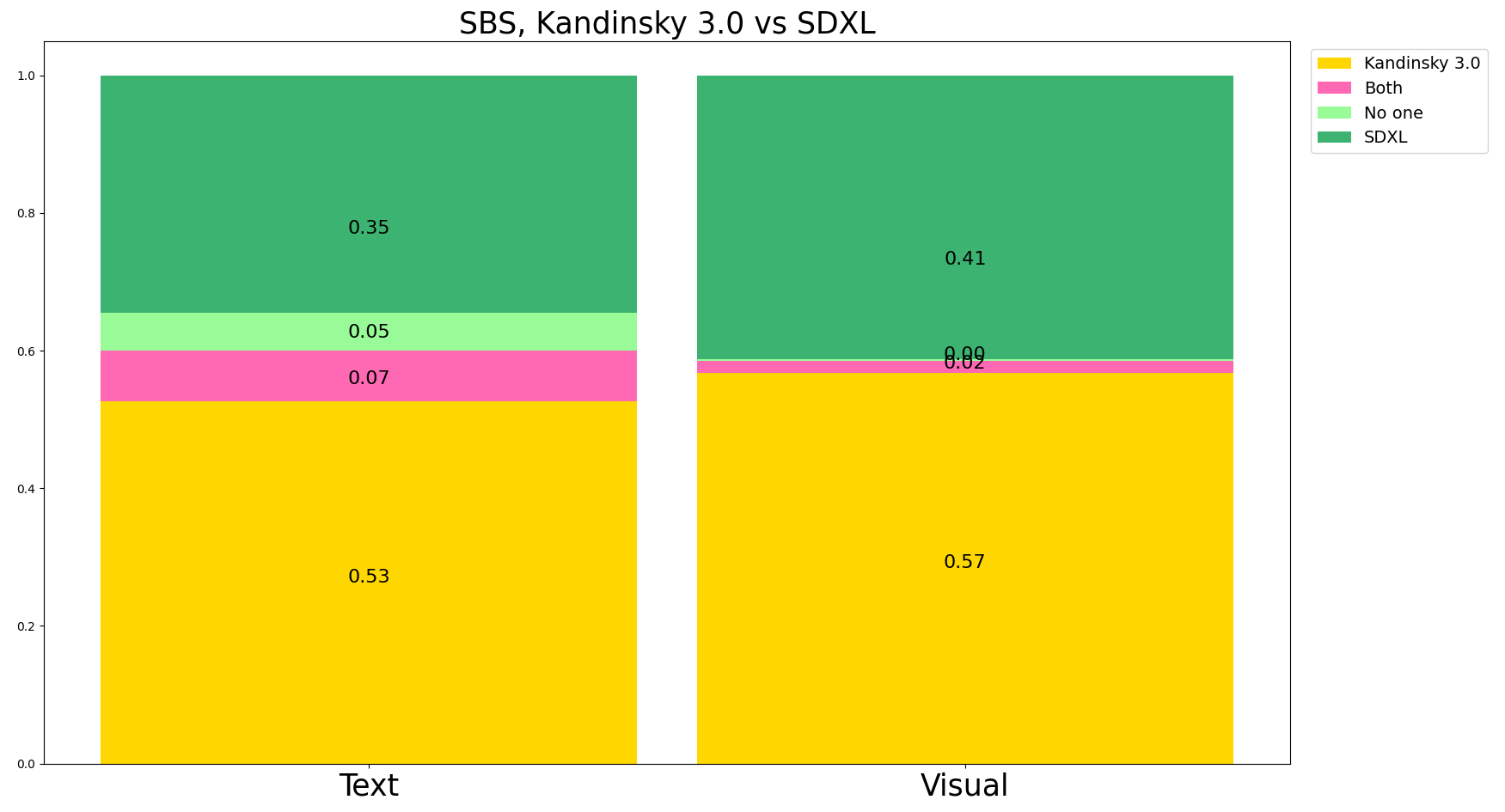}}
    \caption{\centering{Overall results of side-by-side human comparison between Kandinsky 3.0 and SDXL}.}
    \label{fig:kandy3_sdxl_all}
\end{figure}

\begin{figure}[H]
    \center{\includegraphics[bb=0 0 1000 570, scale=0.44
    ]{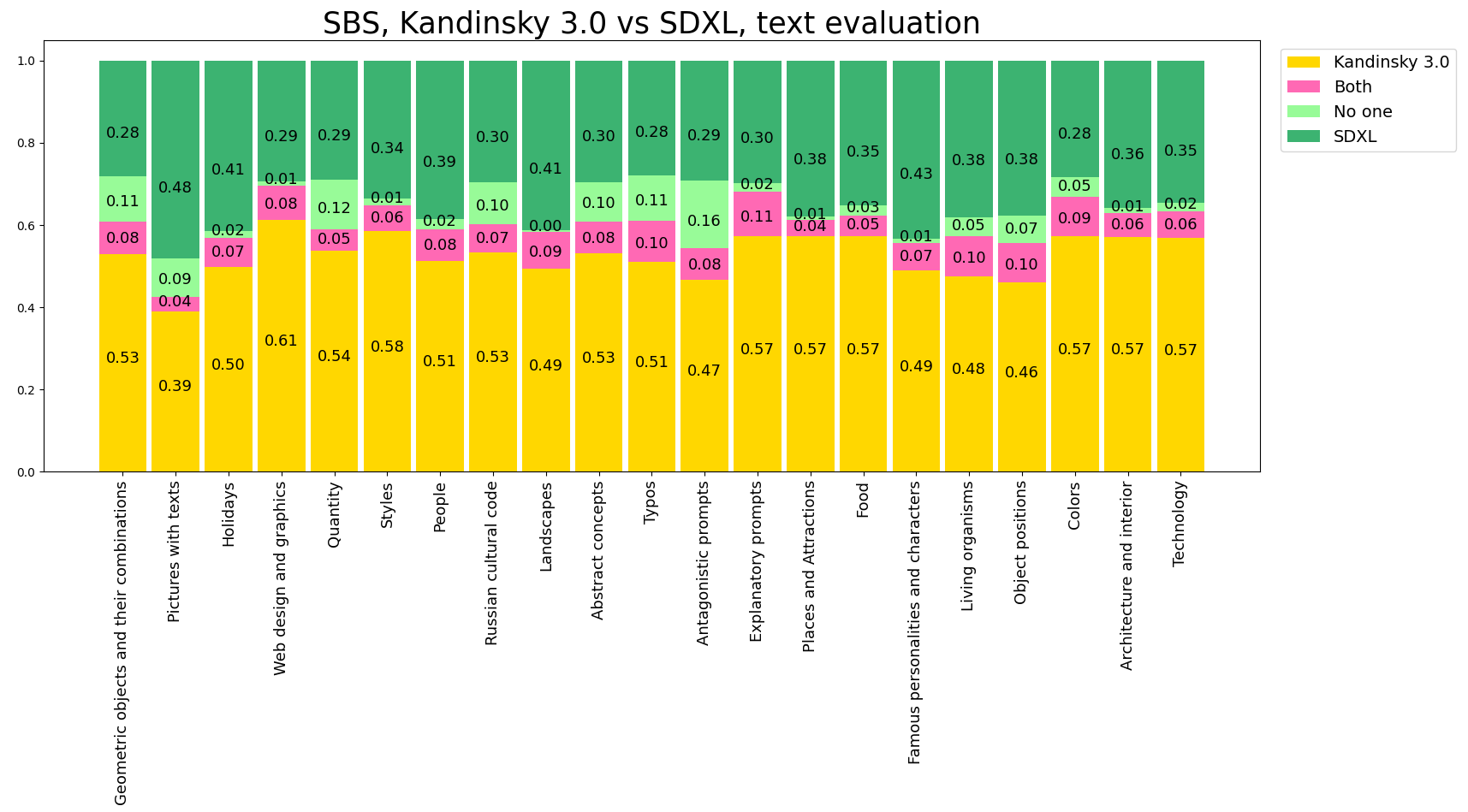}}
    \caption{\centering{Results of side-by-side human comparison between Kandinsky 3.0 and SDXL for \textbf{text comprehension}.}}
    \label{fig:kandy3_sdxl_text}
\end{figure}

\begin{figure}[H]
    \center{\includegraphics[bb=0 0 1000 570, scale=0.44
    ]{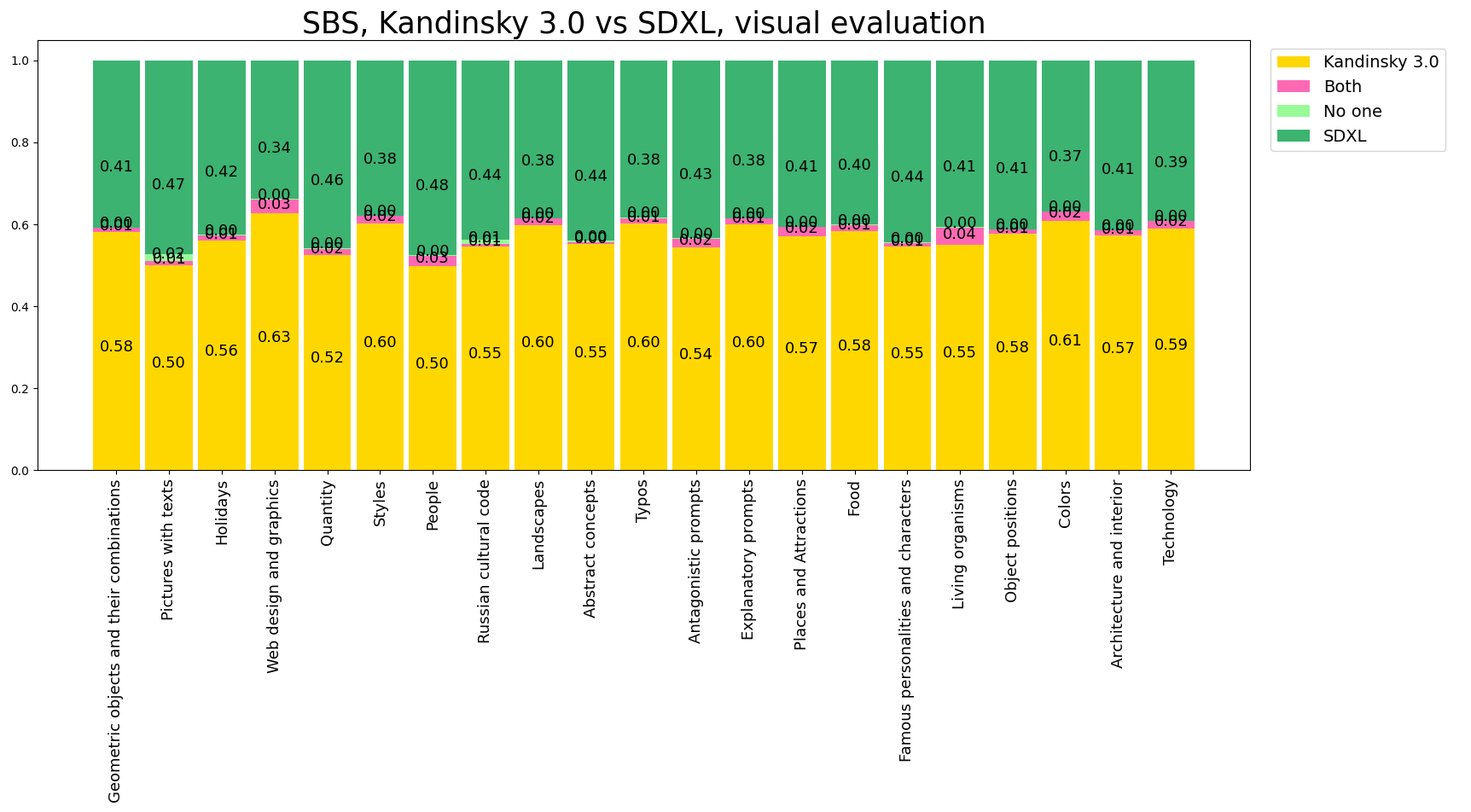}}
    \caption{\centering{{Results of side-by-side human comparison between Kandinsky 3.0 and SDXL for \textbf{visual quality}.}}}
    \label{fig:kandy3_sdxl_visual}
\end{figure}

\subsection*{Human evaluation results for distillation}\label{sec:distillation-sbs-appendix}

\begin{figure}[H]
    \center{\includegraphics[bb=0 0 1000 570, scale=0.44
    ]{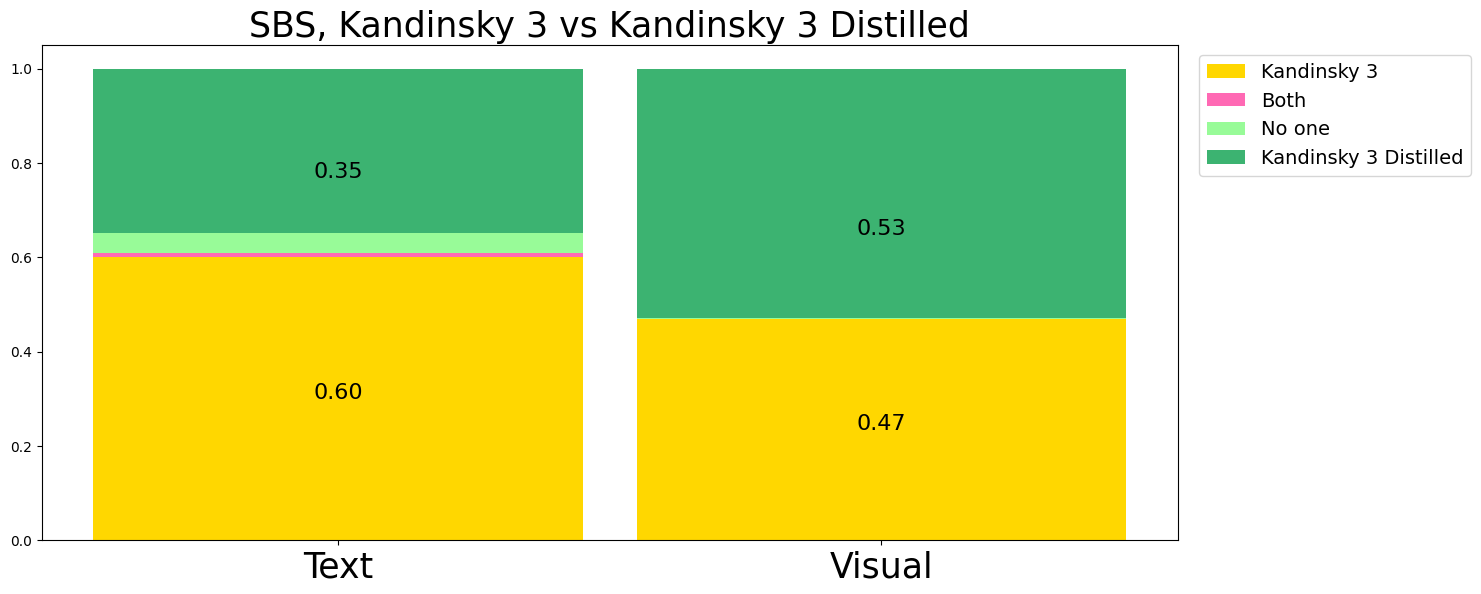}}
    \caption{\centering{Overall results of side-by-side human comparison between Kandinsky 3.0 and Kandinsky 3.0 Distilled (Kandinsky 3.1).}}
    \label{fig:kandy3_distilled_all}
\end{figure}

\begin{figure}[H]
    \center{\includegraphics[bb=0 0 1000 650, scale=0.4
    ]{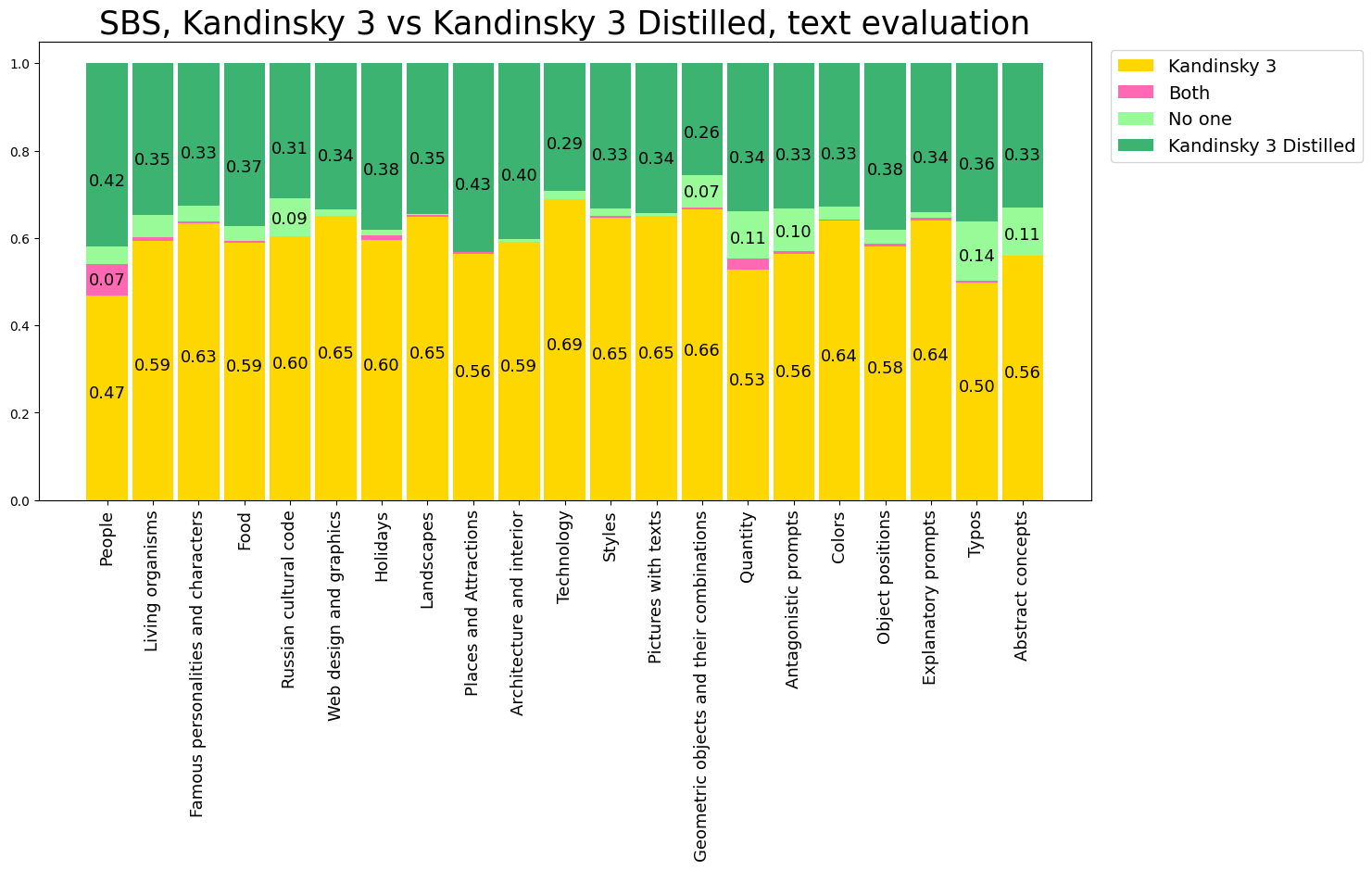}}
    \caption{\centering{Results of side-by-side human comparison between Kandinsky 3.0 and \newline Kandinsky 3.0 Distilled (Kandinsky 3.1) for \textbf{text comprehension}.}}
    \label{fig:kandy3_sdistilled_text}
\end{figure}

\begin{figure}[H]
    \center{\includegraphics[bb=0 0 1000 650, scale=0.4
    ]{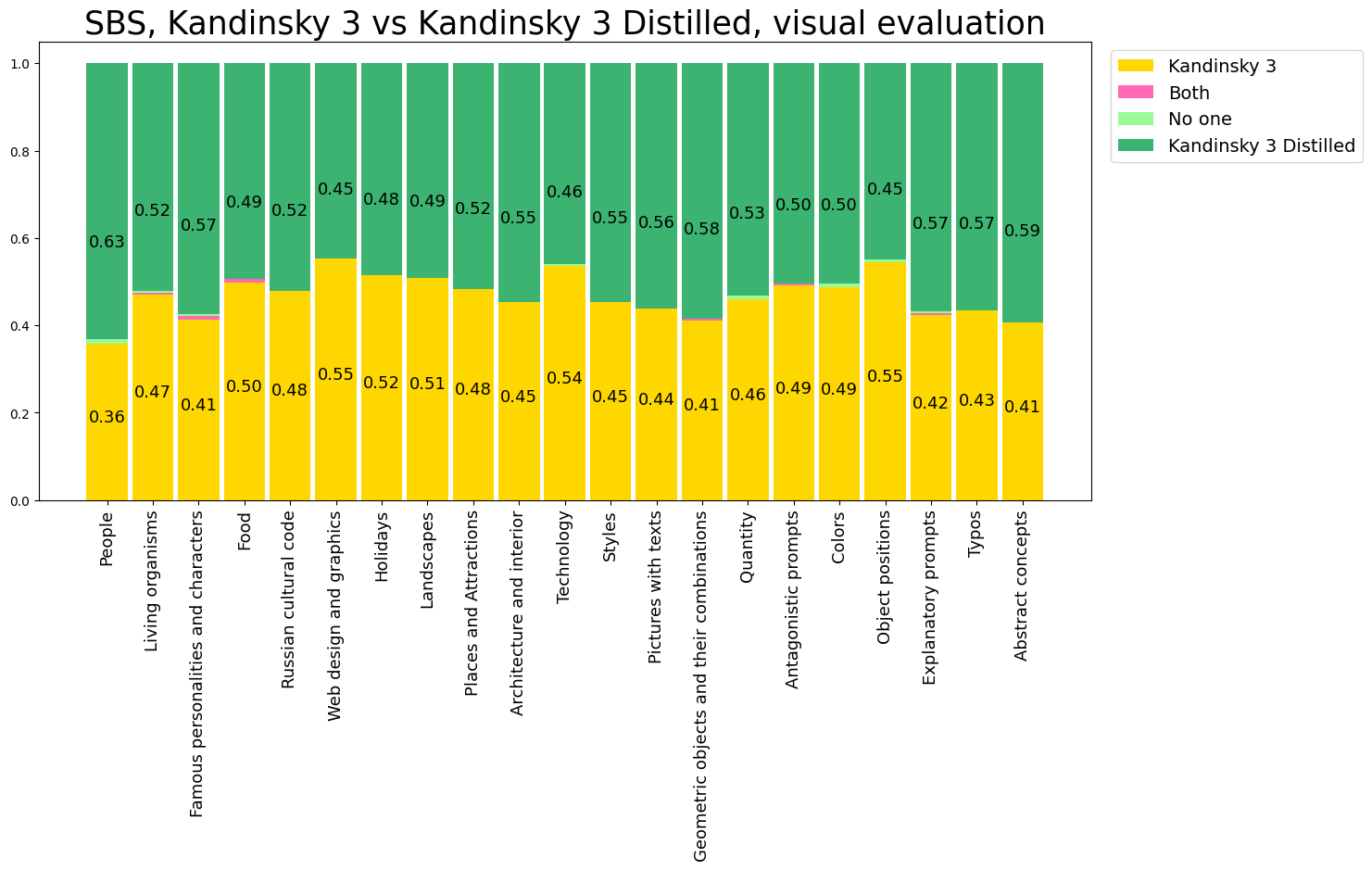}}
    \caption{\centering{{Results of side-by-side human comparison between Kandinsky 3.0 and \newline Kandinsky 3.0 Distilled (Kandinsky 3.1) for \textbf{visual quality}.}}}
    \label{fig:kandy3_distilled_visual}
\end{figure}

\newpage

\subsection*{Human evaluation results for prompt beautification}\label{sec:prompt-beautification-sbs-appendix}

\begin{figure}[H]
    \center{\includegraphics[bb=0 0 1000 450, scale=0.44
    ]{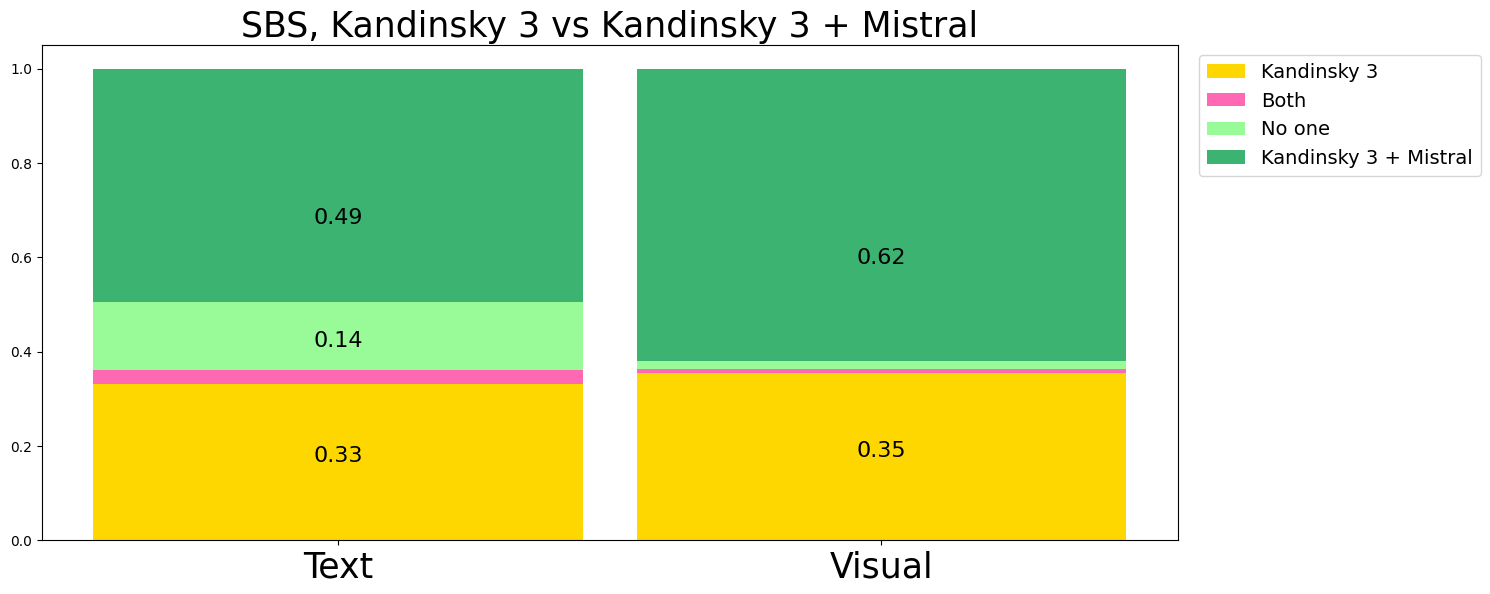}}
    \caption{\centering{Overall results of side-by-side human comparison for Kandinsky 3.0 with and without prompt beautification.}}
    \label{fig:kandy3_prompt_all}
\end{figure}

\begin{figure}[H]
    \center{\includegraphics[bb=0 0 1000 650, scale=0.4
    ]{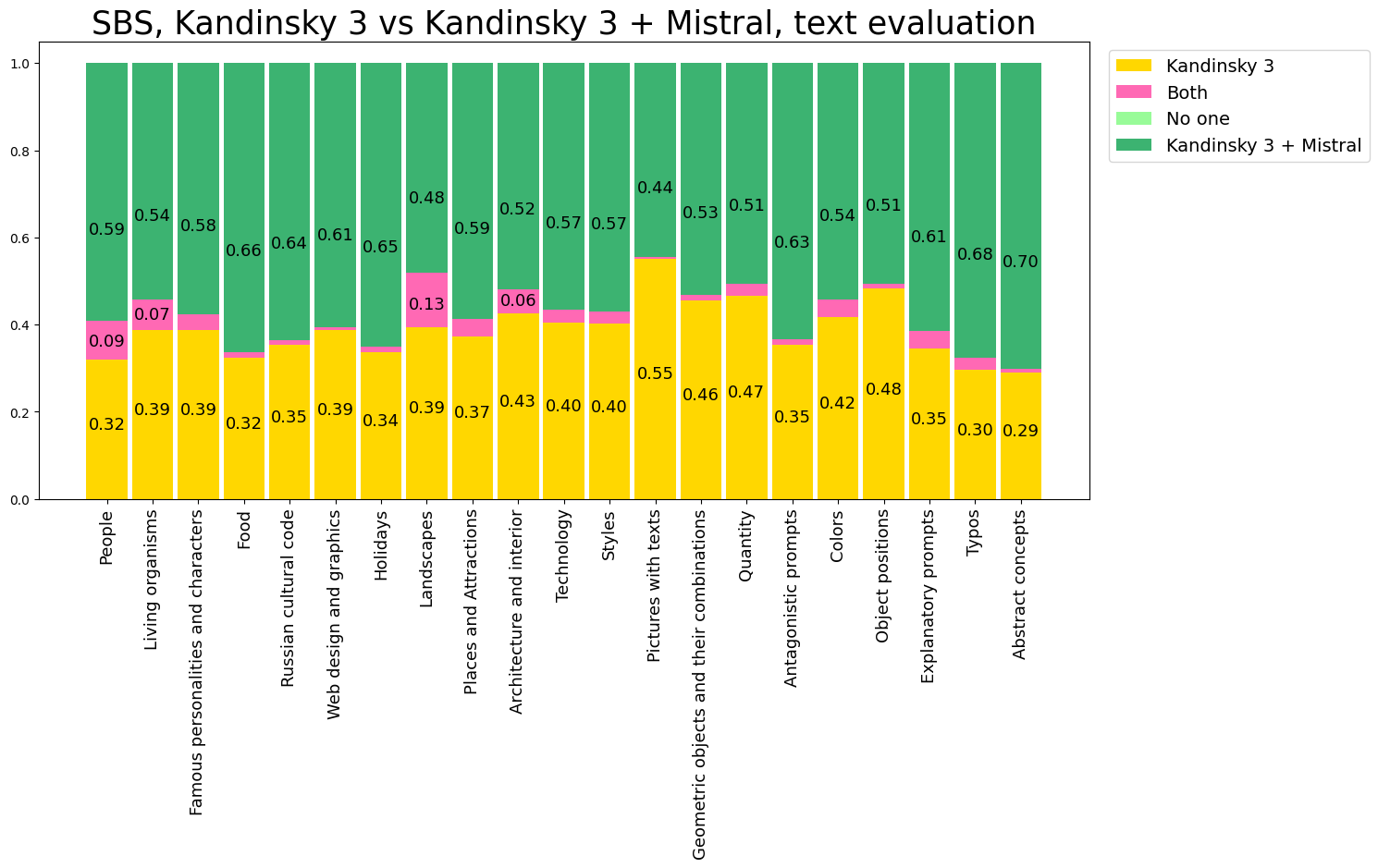}}
    \caption{\centering{Overall results of side-by-side human comparison for Kandinsky 3.0 with and without prompt beautification for \textbf{text comprehension}.}}
    \label{fig:kandy3_prompt_text}
\end{figure}

\newpage

\begin{figure}[H]
    \center{\includegraphics[bb=0 0 1000 650, scale=0.4
    ]{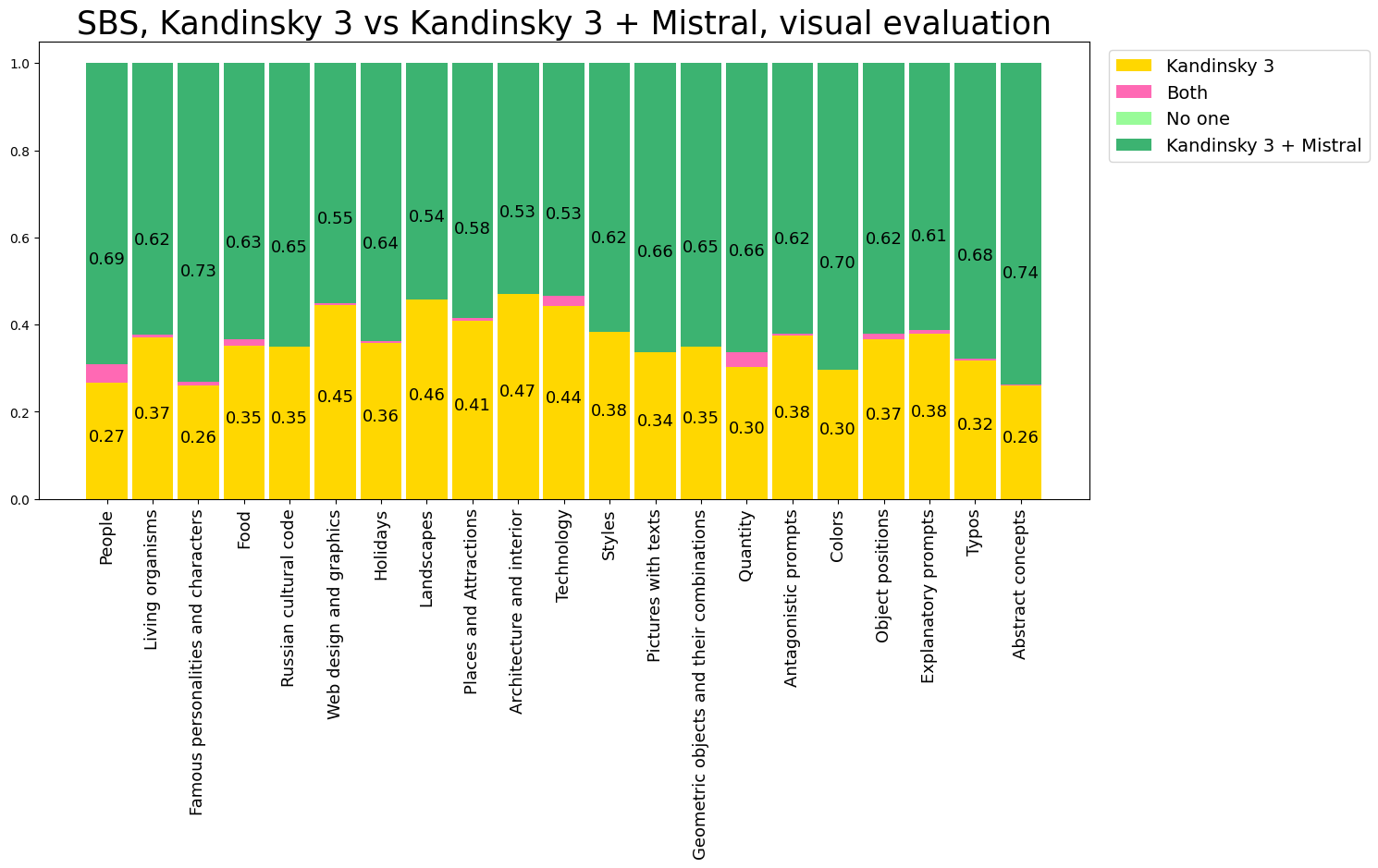}}
    \caption{\centering{Overall results of side-by-side human comparison for Kandinsky 3.0 with and without prompt beautification for \textbf{visual quality}.}}
    \label{fig:kandy3_prompt_visual}
\end{figure}

\begin{figure}[H]
    \center{\includegraphics[bb=0 0 1000 450, scale=0.44
    ]{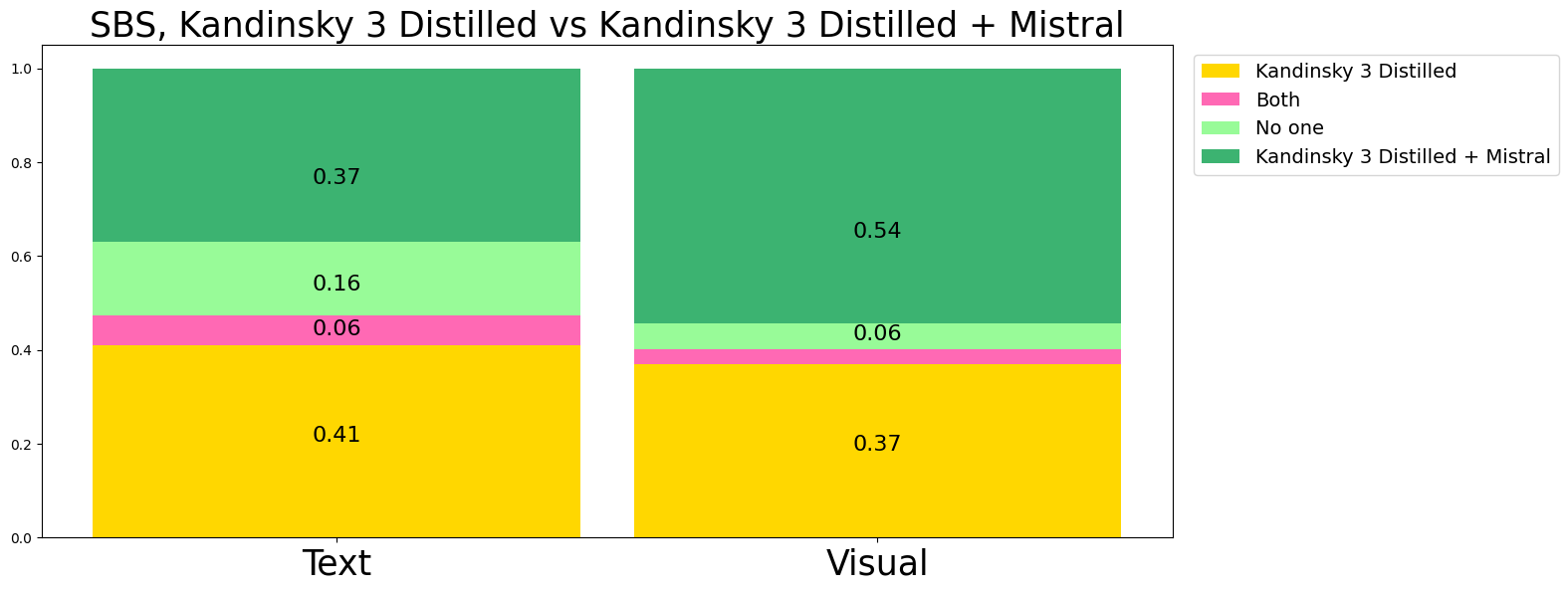}}
    \caption{\centering{Overall results of side-by-side human comparison for Kandinsky 3.0 Distilled (Kandinsky 3.1) with and without prompt beautification.}}
    \label{fig:kandy3_distilled_prompt_all}
\end{figure}

\begin{figure}[H]
    \center{\includegraphics[bb=0 0 1000 650, scale=0.4
    ]{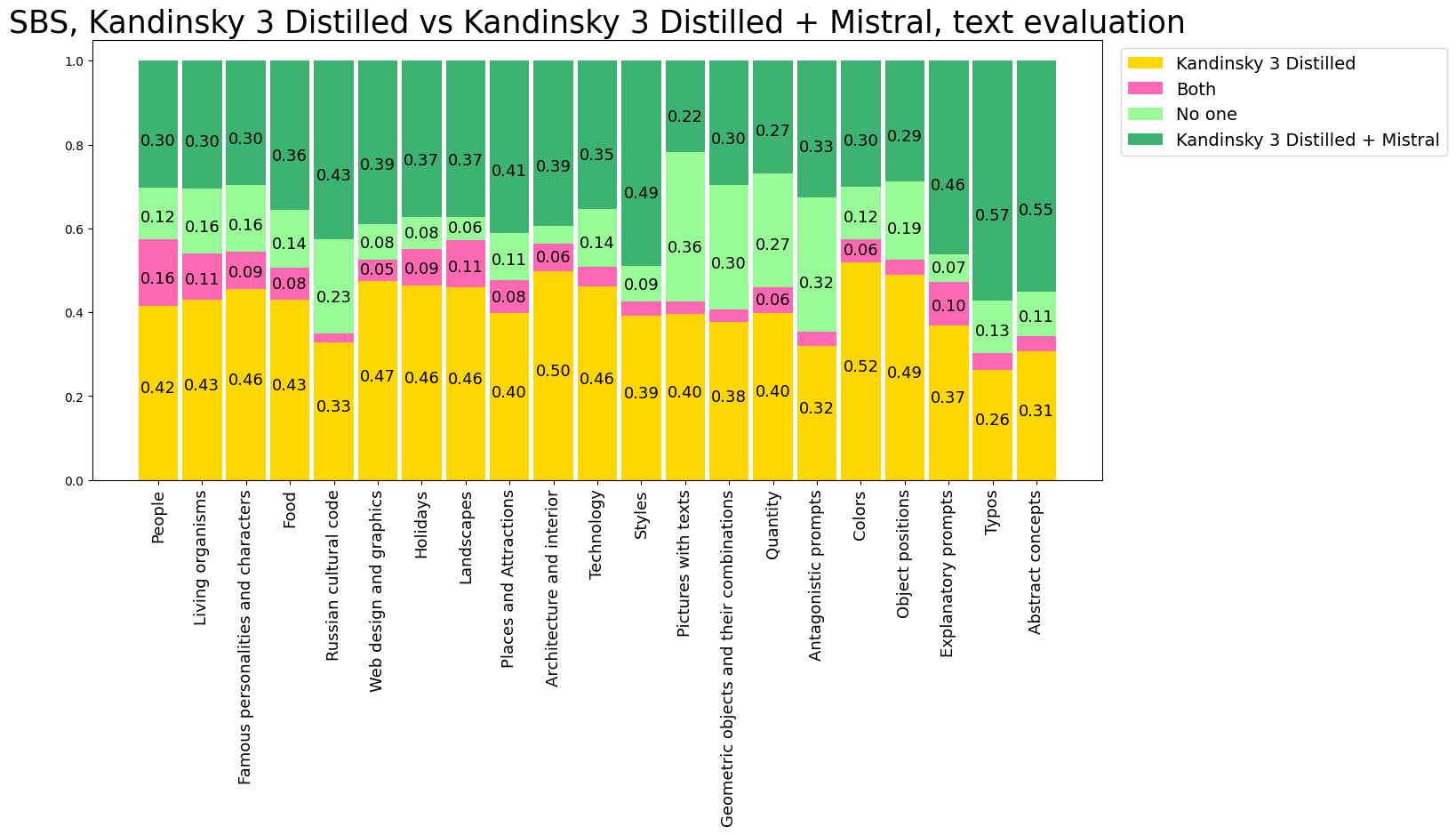}}
    \caption{\centering{Overall results of side-by-side human comparison for Kandinsky 3.0 Distilled (Kandinsky 3.1) with and without prompt beautification for \textbf{text comprehension}.}}
    \label{fig:kandy3_distilled_prompt_text}
\end{figure}

\begin{figure}[H]
    \center{\includegraphics[bb=0 0 1000 650, scale=0.4
    ]{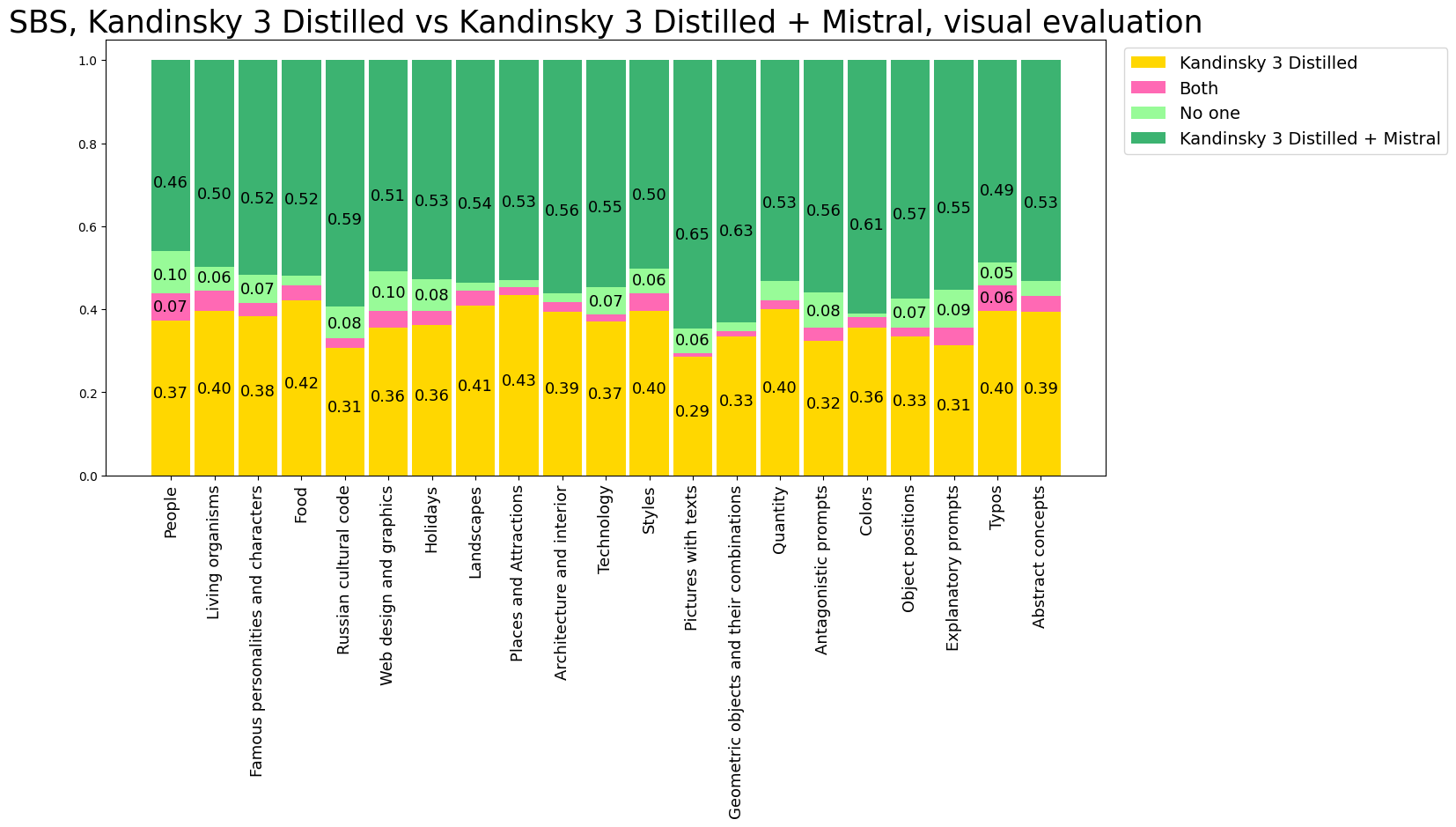}}
    \caption{\centering{Overall results of side-by-side human comparison for Kandinsky 3.0 Distilled (Kandinsky 3.1) with and without prompt beautification for \textbf{visual quality}.}}
    \label{fig:kandy3_distilled_prompt_visual}
\end{figure}

\end{document}